\documentclass[10pt,twocolumn,letterpaper]{article}

\usepackage{iccv}              
\usepackage[accsupp]{axessibility} 
\usepackage{bm}
\usepackage{epsfig}
\usepackage{graphicx}
\usepackage{amsmath}
\usepackage{booktabs}
\usepackage{multirow}
\usepackage{graphics} 
\usepackage{epsfig} 
\usepackage{graphicx}
\usepackage{amsmath}
\usepackage{amssymb}
\usepackage{booktabs}
\usepackage{amsfonts}
\usepackage{graphicx}
\usepackage{pifont}
\usepackage{lipsum}
\usepackage{bbding}
\usepackage{pifont}
\usepackage{wasysym}
\usepackage{arydshln} 
\usepackage{nicematrix}
\usepackage{pifont}
\newcommand{\ym}[1]{{\color{red}#1}}
\definecolor{iccvblue}{rgb}{0.21,0.49,0.74}
\definecolor{cvprblue}{rgb}{0.21,0.49,0.74}
\definecolor{liym_purple}{HTML}{7030A0}
\definecolor{liym_yellow}{RGB}{255,239,213}
\definecolor{arrow_orange}{HTML}{D47C3A}
\definecolor{arrow_blue}{HTML}{5F7FB3}
\definecolor{strength_green}{HTML}{548235}
\definecolor{weakness_orange}{HTML}{C55A11}
\usepackage[pagebackref,breaklinks,colorlinks,allcolors=iccvblue]{hyperref}

\title{TechCoach: Towards Technical-Point-Aware Descriptive Action Coaching}

\author{
    Yuan-Ming Li$^{1,*}$ \hspace{3mm} An-Lan Wang$^{1,*}$ \hspace{3mm} Kun-Yu Lin$^{1}$ \hspace{3mm}  Yu-Ming Tang$^{1}$ \hspace{3mm}  Ling-An Zeng$^{1}$ \\
    Jian-Fang Hu$^{1}$ \hspace{3mm} Wei-Shi Zheng$^{1,\dagger}$ \\
    $^1$iSEE, Sun Yat-sen University \hspace{3mm}\\
    {\tt\small \{liym266, wanganlan\}@mail2.sysu.edu.cn; wszheng@ieee.org }
}

\begin{document}
\maketitle
\begin{abstract}
To guide a learner in mastering action skills, it is crucial for a coach to 1) reason through the learner's action execution and technical points (TechPoints), and 2) provide detailed, comprehensible feedback on what is done well and what can be improved.
However, existing score-based action assessment methods are still far from reaching this practical scenario.
To bridge this gap, we investigate a new task termed Descriptive Action Coaching (DescCoach) which requires the model to provide detailed commentary on what is done well and what can be improved beyond a simple quality score for action execution. 
To this end, we first build a new dataset named EE4D-DescCoach. 
Through an automatic annotation pipeline, our dataset goes beyond the existing action assessment datasets by providing detailed TechPoint-level commentary. 
Furthermore, we propose TechCoach, a new framework that explicitly incorporates TechPoint-level reasoning into the DescCoach process. 
The central to our method lies in the Context-aware TechPoint Reasoner, which enables TechCoach to learn TechPoint-related quality representation by querying visual context under the supervision of TechPoint-level coaching commentary. 
By leveraging the visual context and the TechPoint-related quality representation, a unified TechPoint-aware Action Assessor is then employed to provide the overall coaching commentary together with the quality score.
Combining all of these, we establish a new benchmark for DescCoach and evaluate the effectiveness of our method through extensive experiments. 
The data and code will be made publicly available.
    
\end{abstract}
{\let\thefootnote\relax\footnotetext{
\scriptsize 
{$\dagger$}: Corresponding author. *: Equal contributions.
}}
\section{Introduction}
\begin{figure*}[t]
    \centering
    \includegraphics[width=0.8\linewidth]{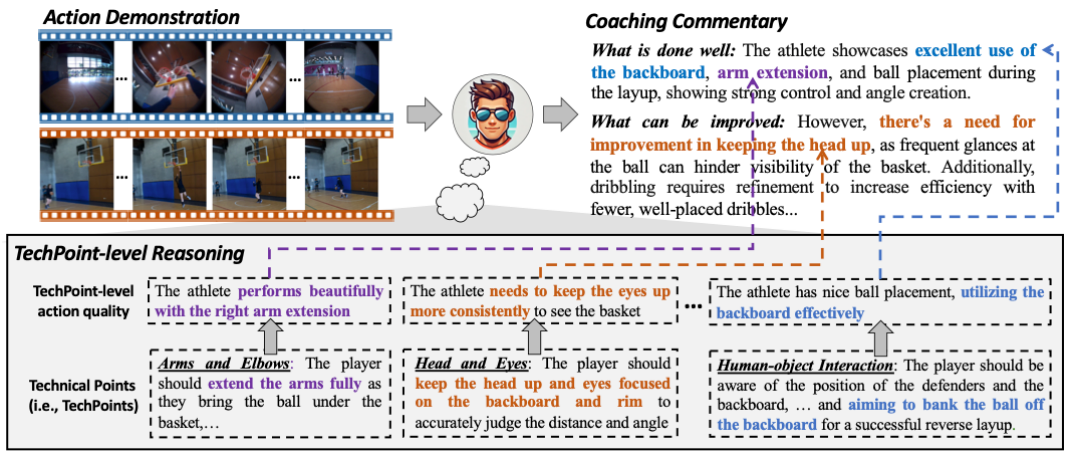}
    \vspace{-0.3cm}
    \caption{\textbf{An illustration of real-world action coaching.} \textbf{Upper}: Given an action demonstration, a human coach will provide coaching comments on \emph{what is done well} and \emph{what can be improved}. 
    \textbf{Lower}: To provide precise feedback, the coach incorporates TechPoint-level Reasoning into the action coaching process, i.e., keep the technical points (\eg, \emph{The player should \textcolor{liym_purple}{\textbf{extend the arms fully as they bring the ball under the basket}}}) in mind and determine the technical-point-level action quality (\eg, \emph{The athlete \textcolor{liym_purple}{\textbf{performs beautifully with the right arm extension}}}) of the action. Such a reasoning process motivates us to propose our TechCoach to address Descriptive Action Coaching.}
    \label{fig:comparison_paradigm}
    \vspace{-0.4cm}
\end{figure*}

Understanding how well an action is performed, also known as Action Quality Assessment (AQA), has recently attached growing attention due to its potential applications \cite{wang2021survey,finediving,parmar2021piano,ding2023sedskill}. One of the most promising potentials of AQA is to build an AI action coach, guiding an action learner toward gradually mastering the skill.
While impressive progress has been achieved, current AQA methods are still far from real coaching and action-guiding scenarios, especially in their functionality and reasoning process:

\noindent{\emph{\textbf{- Functionality}}}. In practical action-guiding scenarios, a coach must provide detailed and understandable feedback on \emph{what is done well} and \emph{what can be improved} so that the learner can be fully aware of the execution details and master the skills (see the upper part of \cref{fig:comparison_paradigm}).
However, existing works formulate AQA as a score-regression \cite{parmar2017learning,usdl,core} or pairwise-ranking \cite{best,shaojie-tmm,egoexolearn} problem, limiting the applicability of the models to more practical scenarios.

\noindent{\emph{\textbf{- Reasoning process}}}. To provide precise feedback, a coach will keep the technical points (TechPoints) in mind (\eg, \emph{The player should extend the arms fully in Reverse Layup}) and determine the TechPoint-level action quality from action execution (see the lower part of \cref{fig:comparison_paradigm}). However, few existing works have incorporated such a TechPoint-level reasoning process into AQA. While recent works \cite{cpr-coach,egoexo-fitness} propose to address this issue by recognizing the existence of TechPoint-level mistakes, such a paradigm still limits deeper exploration of the connections between the action execution and the TechPoint-level action quality (\ie, detailed strength/weakness on a TechPoint).

To bridge the gap between the field of AQA and real-world action coaching scenarios, 
we investigate a new task termed \emph{Descriptive Action Coaching (DescCoach)}, which requires the model to \textbf{provide detailed commentary on \emph{what is done well} and \emph{what can be improved}, beyond merely assigning a quality score to an action execution}. 
This task is challenging because to provide such a detailed commentary, the model not only needs to understand fine-grained action execution (\eg, \emph{the player looks down when performing Reverse Layup}), but also establish explicit connections between the execution with the action quality (\eg, \emph{it is a weakness to look down at the ball as it will miss the target}).

To facilitate the studies on DescCoach, we construct EE4D-DescCoach,
a new dataset that not only contains various action videos and quality scores but also features {hierarchical detailed coaching commentary on both TechPoint and instance levels}. 
Specifically, we first source action videos from the recently proposed EgoExo4D \cite{egoexo4d} dataset. 
Subsequently, we design an LLM-driven automatic annotation pipeline to progressively obtain the general TechPoints and hierarchical coaching commentary. 
{To our knowledge, no existing AQA dataset provides detailed TechPoint-level coaching commentary}. 
Compared to existing datasets with score-based or binary TechPoint-level annotations \cite{egoexo-fitness,iris}, we believe the TechPoint-level commentary provides more clear and detailed clues for reasoning about action quality and providing instance-level coaching commentary.

The construction of the EE4D-DescCoach dataset allows us to build \textbf{Tech}Point-Aware {D}escriptive {A}ction \textbf{Coach} (TechCoach), a new DescCoach framework which explicitly {incorporates TechPoint-level reasoning into the action coaching process under the supervision of TechPoint-level coaching commentary}, 
and delivers feedback by incorporating the reasoning results and visual context.

To this end, TechCoach includes a Context-aware TechPoint Reasoner, which queries the visual context with general TechPoints to obtain TechPoint-related quality representations. To ensure these representations carry the action quality information, a TechPoint-level alignment loss is proposed to align the representations with the TechPoint-level coaching commentary. After that, a Unified TechPoint-aware Action Assessor (TA2) is employed to regress the action quality score and generate overall coaching commentary. 
Throughout this process, a progressive action coaching attention mask guides TA2 to integrate information starting with the visual context level, followed by the TechPoint level, and ultimately the decision level (\ie, score-prediction and commentary-generation).

Combining the EE4D-DescCoach dataset and TechCoach, we establish a new benchmark to expand the field of AQA. 
Extensive experiments demonstrate that: (1) Our TechCoach achieves state-of-the-art performance among the compared methods, including task-specific models and general MLLMs; (2) The Context-aware TechPoint Reasoner is a core design to enhance action quality perceiving ability and cannot be simply replaced by naive alternative solutions (\eg, replacing the TechPoint-level alignment by classifying whether the execution shows strength or weakness on each TechPoint).

The contributions of this work can be summarized as: 
(1) We introduce EE4D-DescCoach, a new dataset specifically designed for Descriptive Action Coaching.
The hierarchical coaching commentary, particularly at the TechPoint level, establishes EE4D-DescCoach as a unique resource among existing AQA datasets.
(2) We develop TechCoach, a new method that effectively integrates TechPoint-level reasoning into the action coaching process through supervision from TechPoint-level coaching commentary. 
(3) We construct a new benchmark on DescCoach. Experiments not only validate the effectiveness of the proposed approach, but also lay the foundation for future research in this field.

\begin{figure*}[t]
    \centering
    \includegraphics[width=0.90\linewidth]{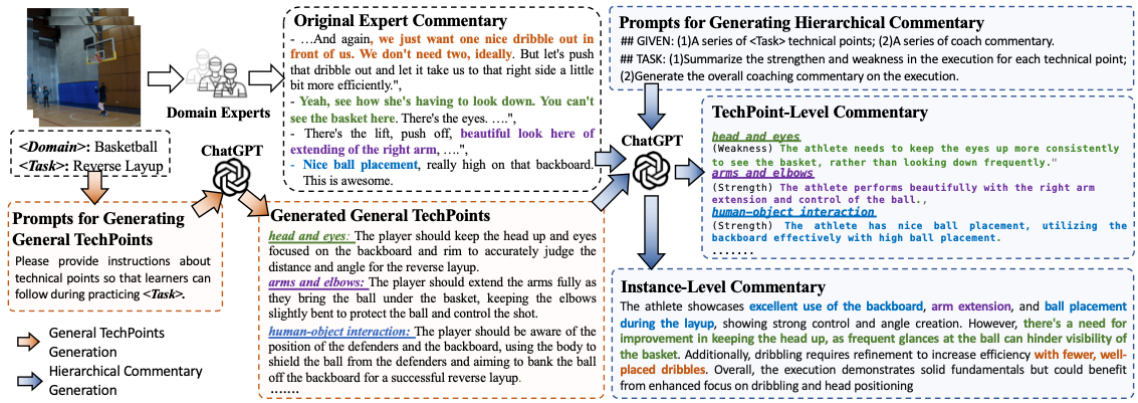}
    \vspace{-0.3cm}
    \caption{\textbf{An overview of the LLM-driven automatic annotation pipeline for the EE4D-DescCoach dataset.} We first prompt the LLM to generate general TechPoints for the given action task (\textbf{\textcolor{arrow_orange}{Orange $\rightarrow$}}). Subsequently, we ask the LLM to summarize the TechPoint-level and instance-level commentary according to the general TechPoints and the original expert commentary (\textbf{\textcolor{arrow_blue}{Blue $\rightarrow$}}). Information about various TechPoints is highlighted in colors. Zoom in for best view.
}
    \label{fig:annot_pipeline}
    \vspace{-0.4cm}
\end{figure*}

\vspace{-0.1cm}
\section{Related Works}
\vspace{-0.1cm}

\noindent{\textbf{Action Quality Assessment (AQA)}} \cite{wang2021survey,liu2024vision,yin2025decade,zhou2024comprehensive} aims at evaluating how well an action is performed.
Most existing works pose AQA as a score regression task, and approach this task in many ways, \eg, introducing new formulations \cite{lgdt,usdl,core}, augmenting human-body representations \cite{jrg,fineparser}, modeling with insufficient or continual data \cite{yun2024semi,li2024continual,zhou2024magr}, exploring temporal structures\cite{finediving,tpt,xu2025vision}, collaborating multi-modal information \cite{parmar2021piano,zeng2024multimodal,xia2023skating}.  
Although impressive progress has been achieved, regressing the final score is still far from real-world action coaching scenarios.
MTL-AQA\cite{mtl-aqa} and NAE\cite{nae} propose to generate commentary in addition to the score, but the commentary comes from live comment, focusing on \emph{action} and \emph{score} rather than skill guiding and coaching.
NS-AQA\cite{okamoto2024hierarchical} combines multiple models together with judging rules to provide formatted feedback for diving judging, but requires complex manual design on a single task.
Differently, our work focuses on generating free-from commentary on \emph{what is done well} and \emph{what can be improved}, which is more aligned with practical coaching scenarios. To this end, we construct a new dataset that features hierarchical (especially the TechPoint-level) detailed coaching commentary. Comparisons with existing datasets are shown in \cref{tab:comparison_datasets}. Furthermore, we propose a general framework that collaborates TechPoint-level reasoning into the coaching process. 
Different from existing works \cite{iris,rica2,fitness-aqa,egoexo-fitness,cpr-coach} that use binary or score-based supervised pipeline to introduce score rubric or detect body-part level mistakes, TechCoach represents the first approach to learn action quality-related representation under the supervision of TechPoint-level commentary. 
Experiments further validate the effectiveness of the unique design against other naive solutions.

\noindent{\textbf{Video Captioning}}  \cite{videobert,xu2024retrieval,swinbert,pdvc,univl,vid2seq} is a long-standing task that requires a model to generate a language caption with video frames as input.
Early studies focus on designing stronger architectures \cite{swinbert,pdvc,aafaq2019spatio,shi2020learning,yamazaki2023vltint} or utilizing video-language pretraining \cite{videobert,univl,vid2seq,alayrac2022flamingo,seo2022end} to model relationships between video and language.
Recently, Multimodal Large Language Models (MLLMs) \cite{video-llava,internvideo2,videochat2,Video-chatgpt,videochat,internvl} show strong generalization ability across various video-language tasks such as video content description and video question answering. 
However, rare exploration has been conducted on whether these models could provide coaching commentary about \emph{what is done well} and \emph{what can be improved}.
In this work, we provide a new testbed for today's video-language models and reveal that they still struggle to address such a practical scenario. Besides, we propose a new TechPoint-guided coaching framework and evaluate its effectiveness through extensive experiments.

\vspace{-0.1cm}
\section{EE4D-DescCoach Dataset}
\vspace{-0.1cm}
To enable studies on Descriptive Action Coaching, we establish a new dataset termed EE4D-DescCoach.
EE4D-DescCoach not only contains various action videos and quality scores but also features {hierarchical detailed coaching commentary on both technical-point (TechPoint) level and instance level}.
We will introduce the data source in \cref{sec:data_source}, the annotation pipeline in \cref{sec:annotation_pipeline}, and provide comparisons with existing datasets in \cref{sec:comparison_exesting_datasets}.

\vspace{-0.1cm}
\subsection{Data Source}
\label{sec:data_source}
\vspace{-0.1cm}
We source data from the recently proposed EgoExo4D \cite{egoexo4d} dataset.
In addition to ego-exo action videos, EgoExo4D also contains execution rating (1 to 10) and time-aligned free-form spoken commentary (e.g. ``\emph{Yeah, see how she's having to look down. You can't see the basket here.}") by domain experts. Such characteristics make the EgoExo4D dataset a great starting point for studying action coaching.

In our work, we focus on physical actions and select takes from EgoExo4D including the scenarios of \emph{basketball}, \emph{soccer}, and \emph{bouldering}. After that, we segment each take into single instances with a fixed-size (8-second) window and filter those instances without meaningful commentary. By doing so, we obtain 4843 unique video instances, spanning about 10.8 hours. 

\begin{table*}[t]
    \scriptsize
    \centering
    \scriptsize
    \resizebox{0.90\linewidth}{!}{
        \begin{NiceTabular}{l|c|ccc|ccc|ccc}
        \toprule
        \multirow{2}{*}{\textbf{Datasets}} & \multirow{2}{*}{\textbf{Public.}} & \textbf{Data}    & \textbf{Unique}  & \textbf{Unique}   & \multicolumn{3}{c|}{\textbf{TechPoint Level Judgment}} & \multicolumn{3}{c}{\textbf{Instance Level Judgment}}        \\ 
        & & \textbf{Sources} & \textbf{Instances} & \textbf{Hours}  & Exists & Type              & Num              & Score                & Commentary & Coach-Score                          \\
        \midrule
        \multicolumn{10}{l}{\emph{\textbf{Datasets w/o Language Annotations}}} \\
        MIT-Skate \cite{mit-skate} & ECCV14 & Olympics & 150 & 7.3h & {$\times$} & - & - & \checkmark & {$\times$} & - \\
        AQA-7 \cite{aqa7} & WACV19 & Olympics & 1189 & 1.62h &{$\times$} & - & - & \checkmark & {$\times$} & - \\
        Rhythmic Gymnastics \cite{zeng2020hybrid} & ACMMM20 & Olympics & 1000 & 23.4h &{$\times$} & - & - & \checkmark & {$\times$} & -\\
        FineDiving \cite{finediving}& CVPR22 & Olympics & 3000 & 3.5h &{$\times$} & - & - & \checkmark & {$\times$} & - \\
        LOGO \cite{logo} & CVPR23 & Olympics & 200 & 11.4h &{$\times$} & - & - & \checkmark & {$\times$} & - \\
        Skate-IRIS \cite{iris} & ACMIUI23 & MIT-Skate \cite{mit-skate} & 150 & 7.3h & \checkmark & Score & 1050 & \checkmark & {$\times$} & - \\
        CPR-Coach \cite{cpr-coach} & CVPR24 & Live Saving & 1416 & 7.7h & \checkmark & Binary & 18.4k & $\times$ & $\times$ & -\\
        \midrule
        \multicolumn{10}{l}{\emph{\textbf{Datasets w/ Language Annotations}}}    \\
        MTL-AQA \cite{mtl-aqa} & CVPR19 & Olympics & 1412 & 1.7h &{$\times$} & - & - & \checkmark & \checkmark & 1.94 \\
        {MTL-NAE} \cite{nae} & {CVPR24} & MTL-AQA \cite{mtl-aqa} & {1412} & 1.7h &{$\times$} & {-}  & {-} & \checkmark & \checkmark & {2.94} \\
        EgoExo-Fitness \cite{egoexo-fitness}& ECCV24 & Daily Fitness & 913 & 4.6h &\checkmark & Binary & 7.8k & \checkmark & \checkmark & 2.37\\
        EgoExo4D (Sports) \cite{egoexo4d} & CVPR24 & Daily Sports & 1219 & 16.4h & {$\times$} & - & - & \checkmark & \checkmark & 3.44\\
        \rowcolor{liym_yellow} {EE4D-DescCoach (Ours)} &  & {EgoExo4D \cite{egoexo4d}}  & \textbf{4843} & 10.8h & \textbf{\checkmark}  & \textbf{Commentary}  & \textbf{25.1k}  & \textbf{\checkmark}  & \textbf{\checkmark} & \textbf{4.33} \\ 
                                
        \bottomrule        
        \end{NiceTabular}
    }
    \vspace{-0.3cm}
    \caption{\textbf{Comparison to popular Action Assessment datasets}. Our EE4D-DescCoach dataset is the first dataset that contains hierarchical (especially TechCoach-level) detailed coaching commentary. Coach-Score: A metric that evaluates whether the provided commentary is suited for real-world coaching scenarios (See \cref{sec:comparison_exesting_datasets} for details).}
    \vspace{-0.4cm}
    \label{tab:comparison_datasets}
\end{table*}

\vspace{-0.1cm}
\subsection{Annotation Pipeline}
\label{sec:annotation_pipeline}
\vspace{-0.1cm}
By diving deeper into the expert commentary in EgoExo4D, it can be observed that: 
1) The original expert commentary is highly colloquial and noisy, awaiting summarization before being used to build a coaching model.
2) The expert commentary is highly associated with the general TechPoints,
making it possible to mine the relationships between the general TechPoints and final coaching commentary.

Based on the observations, we propose an LLM-based automatic annotation pipeline to obtain general TechPoints, and hierarchical coaching commentary on both TechPoint level and instance level, which is shown in \cref{fig:annot_pipeline}. We will bring more details as follows.

\vspace{0.1cm}
\noindent{\textbf{- General TechPoints Collection.}}
In the general coaching-learning scenario, a coach will provide several TechPoints in advance so that the skill learner can learn to follow.  
Considering that completing a skilled physical action always requires cooperation across multiple body parts and objects, {we define a TechPoint as a language instruction on one of the seven dimensions, including six body parts (\eg, \emph{head \& eyes}, \emph{arms \& hands}) and the \emph{human-objects interaction}.}

To obtain TechPoints for each skilled action task, we treat the LLM (GPT-4o) as a general coach. As shown in the lower left part of \cref{fig:annot_pipeline}, by providing the task name (\eg, \emph{Basketball Drills - Reverse Layup}) and detailed instructions to the LLM, we are able to obtain the general TechPoints on the previously mentioned seven dimensions (\eg, For the dimension of ``\emph{head \& eyes}'', we have the TechPoint: ``\emph{The player should keep the head up and eyes focused on the backboard and rim to accurately judge the distance and angle for the reverse layup.}'').

\vspace{0.1cm}
\noindent{\textbf{- Hierarchical Coaching Commentary Collection.}}
Subsequently, we further ask the LLM to mine the relationships between the TechPoints and origin expert commentary and provide summarized hierarchical (\ie, TechPoint level and instance level) coaching commentary about \textbf{\emph{what is done well}} and \textbf{\emph{what can be improved}} in the action execution.

As shown in the right part of \cref{fig:annot_pipeline}, in this phase, LLM is asked to finish two tasks: (1) Review the general TechPoints and the original expert commentary, and then summarize the \emph{strength} and \emph{weakness} in the execution corresponding to each TechPoint. (2) Review the original expert commentary and the generated strengths and weaknesses, and then provide the overall commentary on the execution. 

To sum up, with the annotation pipeline, for each unique instance, we construct rich language annotations including the General TechPoints, TechPoint-level Commentary, and Instance-level Commentary.
Moreover, for each instance we also obtain an average rating by averaging all the ratings provided by different experts.

\vspace{0.1cm}
\noindent{\textbf{- Discussions.}} 
Note that in some cases, an execution could reveal both the strength and weakness aspects on one TechPoint, and some may just reveal one of them. Such a characteristic is inherited from EgoExo4D \cite{egoexo4d}.

\vspace{-0.1cm}
\subsection{Comparison with existing datasets}
\vspace{-0.1cm}
\label{sec:comparison_exesting_datasets}
We compare our dataset with related AQA-related datasets. As shown in \cref{tab:comparison_datasets}, 
the detailed TechPoint-level commentary allows us to distinguish our EE4D-DescCoach dataset from the other related datasets \cite{iris,cpr-coach,egoexo-fitness}.
Furthermore, we conduct an evaluation to show the guiding ability of the instance-level commentary given by different datasets. 
Specifically, we prompt the ChatGPT to rate the commentary from 0 to 5 as the Coach-Score. As shown in the last column in \cref{tab:comparison_datasets}, commentary MTL-AQA \cite{mtl-aqa} and MTL-NAE \cite{nae} is not suit for coaching scenario as the original text mainly focuses on the action and score rather than action guiding and skill improvement.
For more details about the prompts, the pre-defined body parts, the examples and quality ensurance, please refer to the Appendix. 

\vspace{-0.1cm}
\section{TechCoach}
\vspace{-0.1cm}

\subsection{Problem Formulation}
\vspace{-0.1cm}
\label{sec:formulation}
Given an action video $v$, the goal of Descriptive Action Coaching is to train a model that takes the video as input and predicts an overall action quality score together with a paragraph of detailed commentary about \emph{what is done well} and \emph{what can be improved}.

\begin{figure*}[t]
    \centering
    \includegraphics[width=0.85\linewidth]{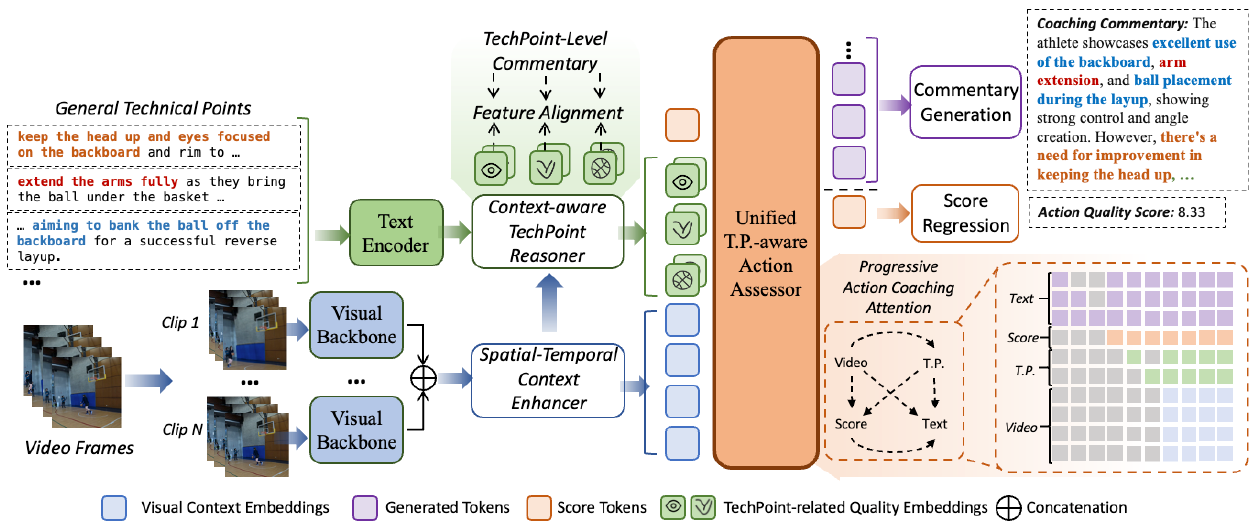}
    \vspace{-0.3cm}
    \caption{\textbf{An overview of the proposed TechCoach.} Given an action video, TechCoach first extracts visual context embeddings with a pre-trained visual backbone and a Spatial-Temporal Context Enhancer. After that, a Context-aware Tech-point Reasoner is adopted to learn Tech-point-related quality embeddings by querying visual context under the supervision of TechPoint-level coaching commentary. 
    Finally, prompted by the visual context and the TechPoint-related quality embeddings, a unified TechPoint-aware Action Assessor is then employed to provide the overall coaching commentary together with the quality score. T.P.: TechPoints. Best viewed in color.}
    \vspace{-0.4cm}
    \label{fig:framework}
\end{figure*}

\begin{figure}[t]
    \centering
    \includegraphics[width=1\linewidth]{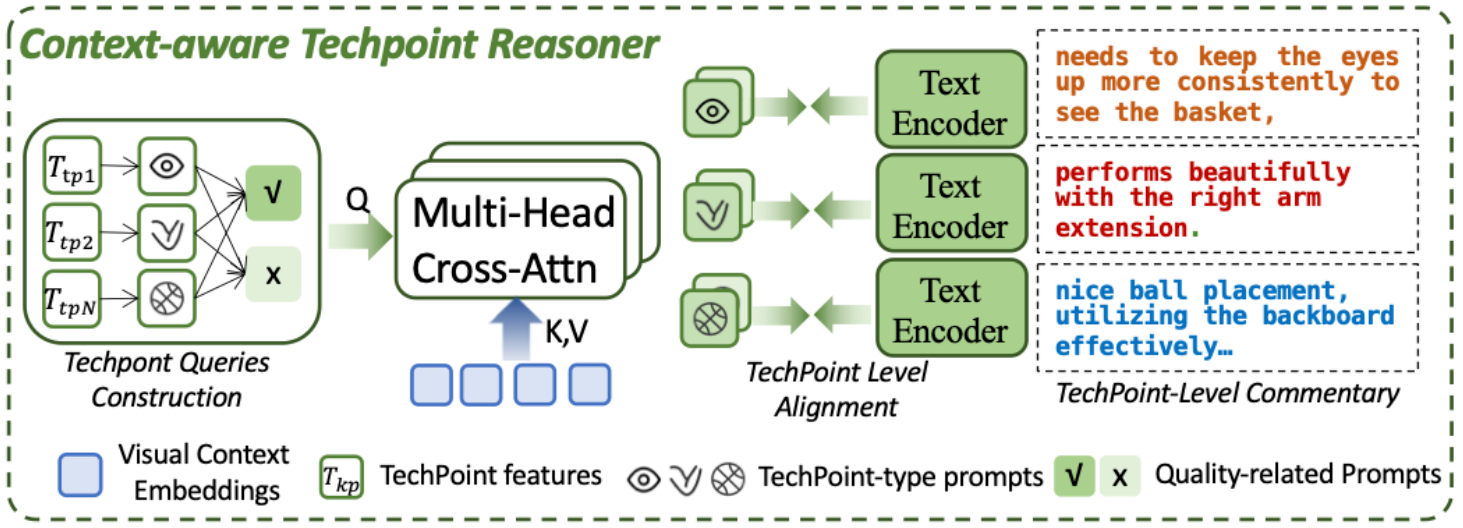}
    \vspace{-0.7cm}
    \caption{\textbf{An overview of the Context-aware TechPoint Reasoner.} Best viewed in color.
    \vspace{-0.4cm}
}
    \label{fig:kp_align}
\end{figure}

\vspace{-0.1cm}
\subsection{Overview}
\vspace{-0.1cm}
\label{sec:pipeline}
An overview of the proposed \textbf{Tech}nical-Point-aware {D}escriptive {A}ction \textbf{Coach} (TechCoach) is shown in \cref{fig:framework}. 
The key idea of TechCoach is to mimic the reasoning process of human coaches by incorporating TechPoint-level reasoning into the action coaching pipeline.
Our framework consists of three parts. Firstly, given an input video $v$, a \textbf{Visual Encoding Module} will extract the visual context embedding.
After that, a \textbf{Context-aware TechPoint Reasoner} is adopted to take the visual context embedding and general technical points as input to perceive the action quality on each TechPoints and provide the TechPoint quality embeddings.
Finally, a \textbf{Unified TechPoint-aware Action Assessor} is utilized to collaborate the visual context and TechPoint-aware quality embeddings, then predict the final quality score and generate coaching commentary.

\vspace{0.1cm}
\noindent{\textbf{Discussions.}} Note that our model is based on the assumption that the general TechPoints are available. We want to argue that this assumption is reasonable because: 1) It is aligned with real-world coaching scenarios that a coach is well-informed about the TechPoints and able to reason through the TechPoints and learner's executions. 2) Obtaining TechPoints is not resource-intensive as we can prompt LLMs (\eg, ChatGPT) to provide the TechPoints.

\vspace{-0.1cm}
\subsection{Visual Encoding}
\vspace{-0.1cm}
Given a video $v$, we first divide the video into non-overlapping
video clips and extract clip-level features with a well-pretrained backbone (\eg, InternVideo2 \cite{internvideo2}). After that, inspired by TimeSformer \cite{timesformer}, we utilize a Transformer-based Spatial-Temporal Context Enhancer to enhance the spatial-temporal context across the clip-level features. This process produces the visual context embeddings $f_v \in R^{T\times H\times W\times D}$ can be obtained, where $T$, $H$, $W$, $D$ indicate the temporal, height, width, and channel size, respectively.
The details about the Spatial-Temporal Context Enhancer are presented in Appendix.

\vspace{-0.1cm}
\subsection{Context-aware TechPoint Reasoner}
\vspace{-0.1cm}
\label{sec:kp_reasoner}
To generate comprehensive coaching commentary for an action execution, an intuitive approach is to directly input the video (and the TechPoint) features into the text generator. However, without explicit modeling, such an approach is unable to effectively perceive the relationship between visual context and the TechPoints. 
To address this, TechCoach adopts a Context-aware TechPoint Reasoner to learn TechPoint-related quality embeddings by querying the video context embeddings under the supervision of the coaching commentary for each TechPoint. An overview of Context-aware TechPoint Reasoner is shown in \cref{fig:kp_align}.

\vspace{0.1cm}
\noindent{\textbf{- TechPoint Queries Construction}}. 
Without loss of generality, given the N general TechPoints corresponding to the input video, we first extract the TechPoint features $f_{tp} \in R^{N\times D}$ with a pre-trained text encoder (\eg, DeBERTa \cite{deberta}) and a linear mapper.
After that, we augment $f_{tp}$ by introducing two types of learnable prompts. Specifically, we use a group of TechPoint-type prompts $f_{tt}\in R^{N\times D}$ to identify which dimension (\eg, head \& eyes, human-object interaction) each TechPoint feature belongs to. Besides, we adopt two quality-related prompts $f_{q}\in R^{2\times D}$ to indicate the aspects of \emph{strength} and \emph{weakness}. Based on the definitions of the learnable prompts, the augmented TechPoint features $f^*_{tp}\in R^{N\times 2 \times D}$ are computed as: 
\vspace{-0.2cm}
\begin{equation}
    f^*_{tp}[i, j] = f_{tp}[i] + f_{tt}[i] + f_{q}[j].
    \vspace{-0.2cm}
\end{equation}
Such an augmentation allows us to inject the TechPoint and quality information into the TechPoint queries. 

\vspace{0.1cm}
\noindent{\textbf{- TechPoint-driven Quality Reasoning.}} Our goal here is to decouple the visual context features corresponding to the TechPoints and quality aspects. To this end, we regard the augmented TechPoint features $f^*_{tp}$ as the \emph{queries} and visual context features $f_v$ as the \emph{keys} and \emph{values}, then utilize layers of multi-head cross attention module \cite{transformer} to obtain the TechPoint-related quality embeddings $f_{tq}\in R^{N\times 2\times D}$. $f_{tq}[i,j]$ indicates the action quality on TechPoint $i$ and aspect $j$ ($1$ for the \emph{strength} and $2$ for the \emph{weakness}).

To ensure $f_{tq}$ carries the quality information related to the action execution on the TechPoints, our idea is to align the TechPoint-related quality embeddings $f_{tq}$ with the features of the corresponding TechPoint-level coaching commentary. To this end, a TechPoint-level alignment loss is adopted, which can be written as:
\vspace{-0.4cm}
\begin{equation}
    L_{Align} = \sum^N_{i=1} \sum^2_{j=1} w_{tc}[i,j] * D(\psi (f_{tq}[i,j]), f_{tc}[i,j]),
    \vspace{-0.4cm}
\end{equation}
where $f_{tc}$ indicates the features of TechPoint-level commentary extracted by the text encoder; $\psi$ is a linear mapper that projects the $f_{tq}$ back to the dimension of the text features; $w_{tc}[i,j]$ is a binary scalar 
to filter those cases where commentary on some aspects is not provided (see Discussions in \cref{sec:annotation_pipeline});
$D(\cdot ,\cdot)$ indicates a distance measurement between two features and we use the L2 Distance by default.

\vspace{-0.1cm}
\subsection{Unified TechPoint-aware Action Assessor}
\vspace{-0.1cm}
After obtaining the visual context embeddings and the TechPoint-related quality embeddings, we employ a Transformer-based Unified TechPoint-aware Action Assessor (TA2) to jointly predict the action quality score and generate the overall coaching commentary. 

Specifically, the input sequence to the TA2 module is constructed by concatenating the following components: (1) The generated text tokens $T_{1:i-1}$; (2) a learnable score-prediction token $f_s$; (3) the flattened $f_{tq}$ as TechPoint tokens; (4) the flattened $f_v$ as video tokens. 
For generating commentary, we follow \cite{swinbert, nae} by employing a Mask-Token Prediction supervision ($L_{MTM}$) during training and using the Next-Token Prediction paradigm during inference.
For score regression, the score-prediction token from the output layer is passed to a fully connected regressor, optimized using a Mean Squared Error (MSE) loss ($L_{MSE}$).

Moreover, to guide the TA2 module in progressively integrating information in a sequence from lower to higher level (\ie, starting with visual context level, then TechPoint level, and finally the decision level), a progressive action coaching attention mask is further adopted on the TA2, which is shown in the lower right part of \cref{fig:framework}. 
Specifically, each TechPoint token independently attends to itself and the video tokens. Besides, the score-prediction token attends to itself and all the TechPoint tokens and video tokens. For commentary generation, TA2 integrates all information by attending to the generated text tokens, the score-prediction token, all the TechPoint tokens, and the video tokens.

To sum up, we train the TechCoach with the following overall loss function:
\vspace{-0.3cm}
\begin{equation}
    L = L_{MTM} + \lambda_1 L_{MSE} + \lambda_2 L_{Align},
    \label{eq:overall_loss}
    \vspace{-0.3cm}
\end{equation}
where $\lambda_1$ and $\lambda_2$ are hyper-parameters.

\begin{table*}[t]
    \centering
    \begin{minipage}{0.38\textwidth}
        \centering
        \scriptsize
        \begin{NiceTabular}{l|c@{\hspace{4pt}}|c@{\hspace{6pt}}c@{\hspace{6pt}}}
            \toprule
            \multirow{2}{*}{\textbf{Methods}} & {\textbf{Visual}} & \multicolumn{2}{c}{\textbf{Score Regression}} \\ 
            & \textbf{Backbone} & $\rho\uparrow$ & RL2$\downarrow$ \\
            \midrule
            \multicolumn{3}{l}{\emph{Models infer with single instance}} \\
            InternVideo2-MLP  & IV2 \cite{internvideo2} &  69.75 & 4.70\\
            USDL \cite{usdl} & IV2 \cite{internvideo2} & 70.81 & 4.51\\
            SwinBERT \cite{swinbert} & IV2 \cite{internvideo2} & 71.37 & 4.59\\
            PGMI \cite{nae} & IV2 \cite{internvideo2} & 70.93 & 4.38\\
            \rowcolor{liym_yellow} TechCoach(Ours) & IV2 \cite{internvideo2} & \textbf{72.46} & \textbf{4.35} \\
            \midrule
            \multicolumn{3}{l}{\emph{Models infer with extra exemplars}} \\
            \textcolor{lightgray}{CoRe \cite{core} \emph{w/ 10 Exem.}} &  \textcolor{lightgray}{IV2 \cite{internvideo2}} & \textcolor{lightgray}{73.17}& \textcolor{lightgray}{4.35} \\
            \textcolor{lightgray}{TPT\cite{tpt} \emph{w/ 10 Exem.}} &  \textcolor{lightgray}{IV2 \cite{internvideo2}} &  \textcolor{lightgray}{73.77} &  \textcolor{lightgray}{4.29} \\
            \bottomrule        
        \end{NiceTabular}
        \subcaption{Comparison with existing methods on score regression.}
    \end{minipage}
    \hspace{0.01\textwidth}
    \begin{minipage}{0.57\textwidth}
        \centering
        \scriptsize
        \begin{NiceTabular}{l@{\hspace{4pt}}|c@{\hspace{4pt}}|c@{\hspace{6pt}}c@{\hspace{6pt}}c@{\hspace{6pt}}c@{\hspace{6pt}}c@{\hspace{6pt}}c@{\hspace{6pt}}}
            \toprule
            \multirow{2}{*}{\textbf{Methods}} & \textbf{\# Params} & \multicolumn{6}{c}{\textbf{Commentary Generation}} \\ 
            & \textbf{(Txt.Gen.)} & B$\uparrow$ & C$\uparrow$ & M$\uparrow$ & BERT$\uparrow$ & GPT-M$\uparrow$ & GPT-Q$\uparrow$\\
            \midrule
            \multicolumn{8}{l}{\emph{General Multi-modal Large Language Models}} \\
            VideoLLaVA-7B \cite{video-llava} & 7B & 23.78 & 1.88 & 10.10 & 55.21 & 16.02 & 14.20\\
            VideoChat2-7B \cite{videochat2} & 7B & 23.18 & 2.98 & 11.59 & 61.37 & 14.09 & 22.28\\
            InternVideo2-\emph{S3}-8B \cite{internvideo2} & 7B & 27.60 & 2.92 & 13.38 & 64.55 & 31.68 & 24.56\\
            InternVL2-8B \cite{internvl2} & 7B & 17.88 & 0.25 & 17.75 & 61.75 & 41.16 & 29.81\\
            InternVL2-76B \cite{internvl2} & 70B & 10.81 & 0.01 & 16.33 & 55.12 & 43.77 & 32.60\\
            \midrule
            \multicolumn{8}{l}{\emph{Task-specific Models}} \\
            SwinBERT \cite{swinbert} & 136M  & 36.11 & 11.70 &  16.01 &66.03 & 46.56 & 34.57\\
            PGMI \cite{nae} & 136M  & 36.78 &  14.00 & 16.24 & 66.42 & 49.02 & 37.04 \\
            \rowcolor{liym_yellow}  TechCoach(Ours) & 136M  & \textbf{37.06} & \textbf{14.62} & \textbf{16.39} & \textbf{66.89} & \textbf{50.44} & \textbf{38.15} \\
            \bottomrule  
        \end{NiceTabular}
        \subcaption{Comparison with existing methods on coaching commentary generation.}
    \end{minipage}
    \vspace{-0.3cm}
    \caption{\textbf{Comparison with existing models}. Our TechCoach outperforms both the general MLLMs and Multi-Task Learning-based AQA models on coaching commentary generation. Besides, TechCoach shows strong score regression abilities with the highest performance among the methods using single instance for predicting the score. ``\emph{w/ 10 Exem.}'': We gray out models that infer with 10 extra exemplars for fair comparisons. ``Txt.Gen.'': Text Generator. ``IN2'': InternVideo2.}
    \vspace{-0.4cm}
    \label{tab:main_comparison}
\end{table*}

\vspace{-0.1cm}
\section{Experiments}
\vspace{-0.1cm}
\subsection{Dataset and Evaluation Metrics}
\vspace{-0.1cm}
\noindent{\textbf{- Dataset}.} To build new benchmark and evaluate the effectiveness of our TechCoach, we separate EE4D-DescCoach into training and evaluation set (with 3769 and 1074 instances) following the official split of EgoExo4D \cite{egoexo4d}.

\vspace{0.1cm}
\noindent{\textbf{- Evaluation Metrics}.}
For score prediction, we use the \emph{Spearman’s rank correlation
coefficient} ($\rho$) and \emph{Relative-L2 Distance} (RL2) as the metrics.
For natural language generation, we adopt NLP metrics, including \emph{BERT} \cite{bertscore},
\emph{BLEU}(B)\cite{bleu}, \emph{METEOR}(M)\cite{meteor} and CIDEr(C) \cite{cider} scores. 
Moreover, for better comparison with open-source MLLMs, we design two more LLM-based Metrics: (1) a \emph{Mention Score}(GPT-M) to evaluate whether the generated commentary mentions the same technical details as in the ground truth; (2) a \emph{Quality Score}(GPT-Q) to evaluate whether the generated commentary shares the same praises or improvement opinions as in the ground truth on those both-mentioned technical details (i.e., action quality perception ability).
Please refer to Appendix for more details.

\vspace{-0.1cm}
\subsection{Implementation Details}
\vspace{-0.1cm}
\label{sec:implementation}
For each video instance, we sample frames and segment them into 32 8-frame clips. Subsequently, we use pretrained InternVideo2 \cite{internvideo2} to extract clip features and adopt average pooling to reduce the spatial and temporal resolutions. obtaining video features with a size of $16(T)\times 8(H)\times 8(W)$.
We use both the ego- and exo-centric (\emph{best-exo} in EgoExo4D) videos as input. We concatenate the multi-view visual features on channel dimension and feed them into a linear projector before going through the Spatial-Temporal Context Enhancer. 
We use similar multi-modal Transformer encoder as in \cite{nae,swinbert} to integrate inputs and generate text.
More implementation details are provided in the Appendix.

\vspace{-0.1cm}
\subsection{Main Results}
\vspace{-0.1cm}
As shown in \cref{tab:main_comparison}, we compare the proposed methods with two branches of baselines: 
(1) \emph{Task-Specific Models}: We select task-specific baselines including USDL \cite{usdl}, CoRe \cite{core}, TPT\cite{tpt}, SwinBERT\cite{swinbert} and PGMI \cite{nae}. For fair comparisons, \textbf{we evaluate all task-specific models by using the same visual backbone (\ie, InternVideo2) to extract features and the same fusion strategy on multi-view features}. 
(2) \emph{General Multi-modal Large Language Models}: We compare popular open-source MLLMs including VideoLLaVA \cite{video-llava}, VideoChat2 \cite{videochat2}, InternVideo2-\emph{S3} \cite{internvideo2}, and InternVL2 \cite{internvl2} under zero-shot evaluation settings. 
We provide more details about the baselines in the Appendix. 

\begin{table}[t]
    \scriptsize
    \centering
    \scriptsize
    \resizebox{1\linewidth}{!}{
        \begin{NiceTabular}{ccc|cc|cccc}
        \toprule
        \multicolumn{3}{c}{\textbf{Variants}}  & \multicolumn{2}{c}{\textbf{Score}} & \multicolumn{4}{c}{\textbf{Commentary}} \\ 
        $L_{MSE}$ & $L_{MTM}$ & $L_{Align}$& $\rho\uparrow$ & RL2$\downarrow$ & B$\uparrow$ & C$\uparrow$ & M$\uparrow$
         & BERT$\uparrow$ \\
        \midrule
        \checkmark && & 70.78 & 4.45 & - & -& -  & - \\
        & \checkmark & & - & -& 36.71 &  13.70 & 16.39 & 66.77 \\
        \checkmark & \checkmark & & 70.23& 4.77&36.42 &  14.26 & 16.09 & 65.72 \\
        \rowcolor{liym_yellow} \checkmark & \checkmark & \checkmark & \textbf{72.46} & \textbf{4.35} & \textbf{37.06} & \textbf{14.62} & \textbf{16.39} & \textbf{66.89}   \\
        \bottomrule        
        \end{NiceTabular}
    }
    \vspace{-0.3cm}
    \caption{\textbf{Ablation studies on training losses}. TechPoint-level alignment is a core design to achieve a better trade-off between score regression and coaching commentary generation.}
    \vspace{-0.4cm}
    \label{tab:ablation_various_losses}
\end{table}

\begin{figure}[t]
    \centering
    \includegraphics[width=1\linewidth]{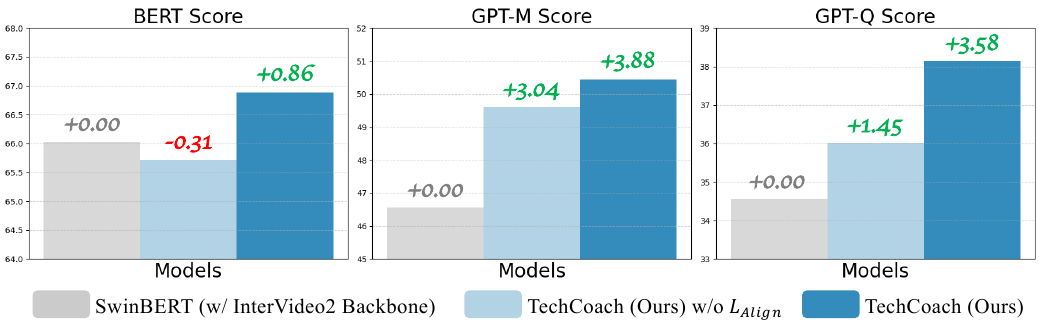}
    \vspace{-0.6cm}
    \caption{\textbf{A further illustration of the impact of TechPoint-level alignment loss.} After adding $L_{Align}$, a much stronger improvement is observed on action quality perception ability (evaluated by GPT-Q Score). Please refer to \cref{sec:ablations} for more analysis. 
}
    \vspace{-0.4cm}
    \label{fig:how_tp_align_work}
\end{figure}

\vspace{0.1cm}
\noindent{\textbf{- Score Regression:}} Compared with Direct Regression-based (\ie, InternVideo2-MLP and USDL) and Multi-Task Learning-based (\ie, SwinBERT, PGMI) AQA methods, TechCoach achieves the best performance on score regression. Note that CoRe and TPT adopt a Contrastive Regression framework that utilizes 10 extra exemplars during inference, thereby enhancing their score regression capability.

\vspace{0.1cm}
\noindent{\textbf{- Coaching Commentary Generation:}} 
(1) \textbf{For task-specific models}: TechCoach achieves SoTA performance on all metrics by explicitly incorporating TechPoint-level reasoning into coaching process.
In \cref{sec:ablations}, we dive deeper to the main componnet of our TechCoach (\ie Context-aware TechPoint Reasoner) by conducting extensive ablation studies.
(2) \textbf{For general MLLMs}: Although shown great visual context understanding and reasoning abilities, the evaluated MLLMs \cite{internvideo2,videochat2,video-llava,internvl2} still fall short on the realistic coaching scenarios (under-perform the task-specific models).
The best among them are the InternVL2 family models, who achieve the highest performance on LLM-based metrics. 
Note that scaling up InternVL2 (8B$\to$76B) brings performance improvement on the LLM-based metrics, but there still exists clear performance gap between general MLLMs and task-specific models.

\vspace{0.1cm}
\noindent{\textbf{- More Comparison Results}}. In Appendix, we also provide deeper explorations and analysis on various aspects, including (i) \emph{User study}, (ii) \emph{Adding general TechPoints into the prompts}, (iii) \emph{Abnormal performance of InternVL2 on traditional language metrics}, (iv) \emph{Fine-tuned MLLM}. 

\begin{table}[t]
    \scriptsize
    \centering
    \scriptsize
    \resizebox{1\linewidth}{!}{
        \begin{NiceTabular}{l|cc|cccc}
        \toprule
        \multirow{2}{*}{\textbf{Variants}} & \multicolumn{2}{c|}{\textbf{Score}} & \multicolumn{4}{c}{\textbf{Commentary}}   \\ 
        & $\rho\uparrow$ & RL2$\downarrow$ & B$\uparrow$ & C$\uparrow$ & M$\uparrow$
         & BERT$\uparrow$ \\
        \midrule
        w/o TP-Align & 70.23& 4.77  &36.42 &  14.26 & 16.09 & 65.72 \\
        TP-CLS & 72.13 & 4.45 & 36.78 & 13.69 & 16.19 & 64.69 \\
        \rowcolor{liym_yellow} TP-Align & \textbf{72.46} & \textbf{4.35} & \textbf{37.06} & \textbf{14.62} & \textbf{16.39} & \textbf{66.89}  \\
        \bottomrule        
        \end{NiceTabular}
    }
    \vspace{-0.3cm}
    \caption{\textbf{Ablation studies on the alternative solutions of TechPoint-level alignment}. The results show that the proposed TechPoint-level alignment (TP-Align) cannot be simply replaced by the naive classification-based solution (TP-CLS).}
    \vspace{-0.4cm}
    \label{tab:kp_align_alternative}
\end{table}

\begin{table}[t]
    \scriptsize
    \centering
    \scriptsize
    \resizebox{1\linewidth}{!}{
        \begin{NiceTabular}{ccc|cc|cccc}
        \toprule
        \multicolumn{3}{c}{\textbf{Variants}}& \multicolumn{2}{c}{\textbf{Score}} & \multicolumn{4}{c}{\textbf{Commentary}}   \\ 
        T.P.Text & T.P.Type &  Quality & $\rho\uparrow$ & RL2$\downarrow$ & B$\uparrow$ & C$\uparrow$ & M$\uparrow$
         & BERT$\uparrow$ \\
        \midrule
        & \checkmark & & 70.83 & 4.49& 36.25 & 13.34 & 15.88 & 66.36  \\
        \checkmark &&  & 71.96 & 4.47 & 35.99 & 13.24 & 15.95 & 66.51\\
        \checkmark & \checkmark & & 72.00 & 4.39 & 36.43 & 12.57 & 16.07 & 66.72 \\
        \rowcolor{liym_yellow} \checkmark & \checkmark & \checkmark & \textbf{72.46} & \textbf{4.35}  & \textbf{37.06} & \textbf{14.62} & \textbf{16.39} & \textbf{66.89}  \\
        \bottomrule        
        \end{NiceTabular}
    }
    \vspace{-0.3cm}
    \caption{\textbf{Ablation studies on the impact of TechPoint Features Augmentation}. T.P.Text: TechPoint text features. T.P.Type: TechPoint-type prompts. Quality: quality-related prompts.}
    \vspace{-0.4cm}
    \label{tab:ablation_kp_aug}
\end{table}

\begin{figure*}[t]
    \centering
    \includegraphics[width=0.85\linewidth]{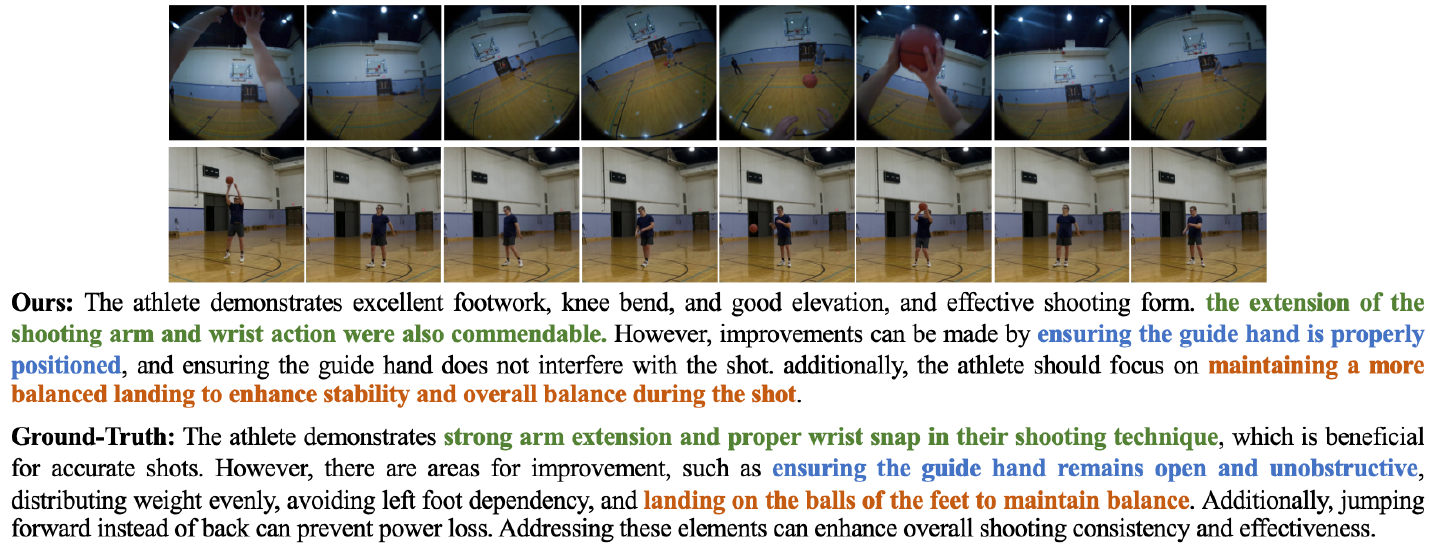}
    \vspace{-0.3cm}
    \caption{\textbf{Qualitative results on the generated coaching commentary.} Our TechCoach generates precise and detailed coaching commentary on \textbf{\emph{what is done well}} and \textbf{\emph{what can be improved}} from the given action videos. Correctly matched parts are highlighted in colors.
}
    \vspace{-0.4cm}
    \label{fig:visualizations}
\end{figure*}

\vspace{-0.2cm}
\subsection{Ablation Studies}
\label{sec:ablations}
\vspace{-0.1cm}
\noindent{\textbf{- Is TechPoint-level alignment necessary? }}We study this question by gradually ablating the training losses illustrated in \cref{eq:overall_loss} from our full model. The results are shown in \cref{tab:ablation_various_losses}. 
By comparing Row 3 with Row 1-2, it can be clearly observed that the model achieves unstable performance when simply combining the $L_{MSE}$ and $L_{MTM}$.
However, when further combining the TechPoint-level alignment loss ($L_{Align}$), we observe stable performance improvement on all the compared metrics (Row 4). 
Results in \cref{fig:how_tp_align_work} further illustrate how $L_{Align}$ impacts the performance. Starting from SwinBERT, after adding all our designs except $L_{Align}$, model performs much better on mentioning the technical details (3.04 improvement on GPT-M Score) due to the TechPoints inputs, but does not obtain similar improvement on perceiving action quality (only 1.45 improvement on GPT-Q Score). After further adding $L_{Align}$, much stronger improvement is observed on GPT-Q Score. All these results indicate the TechPoint-level alignment is necessary and will bring stronger action quality perception ability.

\vspace{0.1cm}
\noindent{\textbf{- Does TechPoint-level alignment equal to binary classification?}}
The main target of the TechPoint-level alignment (TP-Align) is to ensure the TechPoint-related Quality Embeddings (see \cref{fig:framework} and \cref{sec:kp_reasoner} for more details) carry the quality-related information. 
To achieve this, a simple solution is to perform binary classifications over the TechPoint-related Quality Embeddings (\ie, classify whether the execution shows strengths or weaknesses on each TechPoint), which we note as TP-CLS.
To dive deeper, we conduct a study on the influence of such an alternative solution. As shown in \cref{tab:kp_align_alternative}, by comparing Row 2 with Row 1, we observe such a solution strengthens the quality perception of the model, resulting in performance improvement on score regression metrics. However, without explicit alignment with the TechPoint-level coaching commentary, Bi-CLS falls short in understanding deeper relationships between the action execution and TechPoints, resulting in a clear performance gap between our TechPoint-level alignment, especially on commentary generation.

\vspace{0.1cm}
\noindent{\textbf{- Impacts of the TechPoint Query Augmentation.}}
As discussed in \cref{sec:kp_reasoner}, we construct TechPoint queries by augmenting the TechPoint text features (T.P.Text) with TechPoint-type prompts (T.P.Type) and quality-related prompts (Quality). In \cref{tab:ablation_kp_aug}, we gradually ablate the components from the augmentation, and the performance deterioration shows the contribution of each component.

\vspace{0.1cm}
\noindent{\textbf{- More ablation studies.}} We also provide analysis on other aspects like (i) \emph{Impacts of various types of TechPoint-level alignment loss}; (ii) \emph{Influence of the training videos from different views}. Please refer to the Appendix for more details.

\vspace{-0.1cm}
\subsection{Qualitative Results}
\vspace{-0.1cm}
To further illustrate the effectiveness of the proposed TechCoach, we provide qualitative results on the generated coaching commentary. As shown in \cref{fig:visualizations}, TechCoach is able to understand various technical TechPoints of different actions and provide detailed commentary on \textbf{\emph{what is done well}} (\eg, ``the extension of the shooting arm and wrist action were also commendable'') and \textbf{\emph{what can be improved}} (\eg, ``improvements are needed in core engagement to prevent being pulled from the wall'') from the given action videos. More qualitative results are shown in the Appendix.

\vspace{-0.2cm}
\section{Conclusion}
\vspace{-0.1cm}
We investigate Descriptive Action Coaching, a novel task that aims to provide coaching feedback on \emph{what is done well} and \emph{what can be improved} from an action execution. 
To support this task, we develop an automated pipeline for constructing the EE4D-DescCoach dataset which features clean and detailed coaching commentary on both TechPoint and instance levels. 
The TechPoint-level commentary provides new supervision for incorporating TechPoint-level reasoning into the action coaching process, encouraging us to build TechCoach, a new framework empowered by a Context-aware TechPoint Reasoner.
Besides demonstrating strong performance of the proposed TechCoach, extensive experiments also highlights the effectiveness and the necessity of the proposed Context-aware TechPoint Reasoner.
In Appendix, we further discuss the limitations of our work and outline future research directions, including (i) \emph{Scaling up TechCoach}, (ii) \emph{Hierarchical TechPoints}, and (iii) \emph{Pose-Assisted DescCoach}.
We expect the proposed new task, dataset, and method provide a new perspective for advancing current AQA to a more explainable and practical scenario.
{
    \small
    \bibliographystyle{ieeenat_fullname}
    \bibliography{main}

\begin{thebibliography}{63}
\providecommand{\natexlab}[1]{#1}
\providecommand{\url}[1]{\texttt{#1}}
\expandafter\ifx\csname urlstyle\endcsname\relax
  \providecommand{\doi}[1]{doi: #1}\else
  \providecommand{\doi}{doi: \begingroup \urlstyle{rm}\Url}\fi

\bibitem[Aafaq et~al.(2019)Aafaq, Akhtar, Liu, Gilani, and Mian]{aafaq2019spatio}
Nayyer Aafaq, Naveed Akhtar, Wei Liu, Syed~Zulqarnain Gilani, and Ajmal Mian.
\newblock Spatio-temporal dynamics and semantic attribute enriched visual encoding for video captioning.
\newblock In \emph{Proceedings of the IEEE/CVF conference on computer vision and pattern recognition}, pages 12487--12496, 2019.

\bibitem[Alayrac et~al.(2022)Alayrac, Donahue, Luc, Miech, Barr, Hasson, Lenc, Mensch, Millican, Reynolds, et~al.]{alayrac2022flamingo}
Jean-Baptiste Alayrac, Jeff Donahue, Pauline Luc, Antoine Miech, Iain Barr, Yana Hasson, Karel Lenc, Arthur Mensch, Katherine Millican, Malcolm Reynolds, et~al.
\newblock Flamingo: a visual language model for few-shot learning.
\newblock \emph{Advances in neural information processing systems}, 35:\penalty0 23716--23736, 2022.

\bibitem[Bai et~al.(2022)Bai, Zhou, Zhang, Wang, Ding, Guan, Long, and Wang]{tpt}
Yang Bai, Desen Zhou, Songyang Zhang, Jian Wang, Errui Ding, Yu Guan, Yang Long, and Jingdong Wang.
\newblock Action quality assessment with temporal parsing transformer.
\newblock In \emph{European conference on computer vision}, pages 422--438. Springer, 2022.

\bibitem[Banerjee and Lavie(2005)]{meteor}
Satanjeev Banerjee and Alon Lavie.
\newblock Meteor: An automatic metric for mt evaluation with improved correlation with human judgments.
\newblock In \emph{Proceedings of the acl workshop on intrinsic and extrinsic evaluation measures for machine translation and/or summarization}, pages 65--72, 2005.

\bibitem[Bertasius et~al.(2021)Bertasius, Wang, and Torresani]{timesformer}
Gedas Bertasius, Heng Wang, and Lorenzo Torresani.
\newblock Is space-time attention all you need for video understanding?
\newblock In \emph{ICML}, page~4, 2021.

\bibitem[Chen et~al.(2024{\natexlab{a}})Chen, Wang, Tian, Ye, Gao, Cui, Tong, Hu, Luo, Ma, et~al.]{internvl2}
Zhe Chen, Weiyun Wang, Hao Tian, Shenglong Ye, Zhangwei Gao, Erfei Cui, Wenwen Tong, Kongzhi Hu, Jiapeng Luo, Zheng Ma, et~al.
\newblock How far are we to gpt-4v? closing the gap to commercial multimodal models with open-source suites.
\newblock \emph{arXiv preprint arXiv:2404.16821}, 2024{\natexlab{a}}.

\bibitem[Chen et~al.(2024{\natexlab{b}})Chen, Wu, Wang, Su, Chen, Xing, Zhong, Zhang, Zhu, Lu, et~al.]{internvl}
Zhe Chen, Jiannan Wu, Wenhai Wang, Weijie Su, Guo Chen, Sen Xing, Muyan Zhong, Qinglong Zhang, Xizhou Zhu, Lewei Lu, et~al.
\newblock Internvl: Scaling up vision foundation models and aligning for generic visual-linguistic tasks.
\newblock In \emph{Proceedings of the IEEE/CVF Conference on Computer Vision and Pattern Recognition}, pages 24185--24198, 2024{\natexlab{b}}.

\bibitem[Ding et~al.(2023)Ding, Xu, and Li]{ding2023sedskill}
Xinpeng Ding, Xiaowei Xu, and Xiaomeng Li.
\newblock Sedskill: Surgical events driven method for skill assessment from thoracoscopic surgical videos.
\newblock In \emph{International Conference on Medical Image Computing and Computer-Assisted Intervention}, pages 35--45. Springer, 2023.

\bibitem[Doughty et~al.(2019)Doughty, Mayol-Cuevas, and Damen]{best}
Hazel Doughty, Walterio Mayol-Cuevas, and Dima Damen.
\newblock The pros and cons: Rank-aware temporal attention for skill determination in long videos.
\newblock In \emph{Proceedings of the IEEE/CVF conference on computer vision and pattern recognition}, pages 7862--7871, 2019.

\bibitem[Grauman et~al.(2024)Grauman, Westbury, Torresani, Kitani, Malik, Afouras, Ashutosh, Baiyya, Bansal, Boote, et~al.]{egoexo4d}
Kristen Grauman, Andrew Westbury, Lorenzo Torresani, Kris Kitani, Jitendra Malik, Triantafyllos Afouras, Kumar Ashutosh, Vijay Baiyya, Siddhant Bansal, Bikram Boote, et~al.
\newblock Ego-exo4d: Understanding skilled human activity from first-and third-person perspectives.
\newblock In \emph{Proceedings of the IEEE/CVF Conference on Computer Vision and Pattern Recognition}, pages 19383--19400, 2024.

\bibitem[He et~al.(2020)He, Liu, Gao, and Chen]{deberta}
Pengcheng He, Xiaodong Liu, Jianfeng Gao, and Weizhu Chen.
\newblock Deberta: Decoding-enhanced bert with disentangled attention.
\newblock \emph{arXiv preprint arXiv:2006.03654}, 2020.

\bibitem[Hu et~al.(2022)Hu, Shen, Wallis, Allen-Zhu, Li, Wang, Wang, Chen, et~al.]{lora}
Edward~J Hu, Yelong Shen, Phillip Wallis, Zeyuan Allen-Zhu, Yuanzhi Li, Shean Wang, Lu Wang, Weizhu Chen, et~al.
\newblock Lora: Low-rank adaptation of large language models.
\newblock \emph{ICLR}, 1\penalty0 (2):\penalty0 3, 2022.

\bibitem[Huang et~al.(2024)Huang, Chen, Xu, Zhang, Yang, Pei, Zhang, Dong, Wang, Wang, et~al.]{egoexolearn}
Yifei Huang, Guo Chen, Jilan Xu, Mingfang Zhang, Lijin Yang, Baoqi Pei, Hongjie Zhang, Lu Dong, Yali Wang, Limin Wang, et~al.
\newblock Egoexolearn: A dataset for bridging asynchronous ego-and exo-centric view of procedural activities in real world.
\newblock In \emph{Proceedings of the IEEE/CVF Conference on Computer Vision and Pattern Recognition}, pages 22072--22086, 2024.

\bibitem[Li et~al.(2023)Li, He, Wang, Li, Wang, Luo, Wang, Wang, and Qiao]{videochat}
KunChang Li, Yinan He, Yi Wang, Yizhuo Li, Wenhai Wang, Ping Luo, Yali Wang, Limin Wang, and Yu Qiao.
\newblock Videochat: Chat-centric video understanding.
\newblock \emph{arXiv preprint arXiv:2305.06355}, 2023.

\bibitem[Li et~al.(2024{\natexlab{a}})Li, Wang, He, Li, Wang, Liu, Wang, Xu, Chen, Luo, et~al.]{videochat2}
Kunchang Li, Yali Wang, Yinan He, Yizhuo Li, Yi Wang, Yi Liu, Zun Wang, Jilan Xu, Guo Chen, Ping Luo, et~al.
\newblock Mvbench: A comprehensive multi-modal video understanding benchmark.
\newblock In \emph{Proceedings of the IEEE/CVF Conference on Computer Vision and Pattern Recognition}, pages 22195--22206, 2024{\natexlab{a}}.

\bibitem[Li et~al.(2024{\natexlab{b}})Li, Huang, Wang, Zeng, Meng, and Zheng]{egoexo-fitness}
Yuan-Ming Li, Wei-Jin Huang, An-Lan Wang, Ling-An Zeng, Jing-Ke Meng, and Wei-Shi Zheng.
\newblock Egoexo-fitness: Towards egocentric and exocentric full-body action understanding.
\newblock \emph{European Conference on Computer Vision}, 2024{\natexlab{b}}.

\bibitem[Li et~al.(2024{\natexlab{c}})Li, Zeng, Meng, and Zheng]{li2024continual}
Yuan-Ming Li, Ling-An Zeng, Jing-Ke Meng, and Wei-Shi Zheng.
\newblock Continual action assessment via task-consistent score-discriminative feature distribution modeling.
\newblock \emph{IEEE Transactions on Circuits and Systems for Video Technology}, 2024{\natexlab{c}}.

\bibitem[Lin et~al.(2023)Lin, Ye, Zhu, Cui, Ning, Jin, and Yuan]{video-llava}
Bin Lin, Yang Ye, Bin Zhu, Jiaxi Cui, Munan Ning, Peng Jin, and Li Yuan.
\newblock Video-llava: Learning united visual representation by alignment before projection.
\newblock \emph{arXiv preprint arXiv:2311.10122}, 2023.

\bibitem[Lin et~al.(2022)Lin, Li, Lin, Ahmed, Gan, Liu, Lu, and Wang]{swinbert}
Kevin Lin, Linjie Li, Chung-Ching Lin, Faisal Ahmed, Zhe Gan, Zicheng Liu, Yumao Lu, and Lijuan Wang.
\newblock Swinbert: End-to-end transformers with sparse attention for video captioning.
\newblock In \emph{Proceedings of the IEEE/CVF Conference on Computer Vision and Pattern Recognition}, pages 17949--17958, 2022.

\bibitem[Liu et~al.(2024)Liu, Wang, Stawarz, Li, Fu, and Liu]{liu2024vision}
Jiang Liu, Huasheng Wang, Katarzyna Stawarz, Shiyin Li, Yao Fu, and Hantao Liu.
\newblock Vision-based human action quality assessment: A systematic review.
\newblock \emph{Expert Systems with Applications}, page 125642, 2024.

\bibitem[Loshchilov(2017)]{adamw}
I Loshchilov.
\newblock Decoupled weight decay regularization.
\newblock \emph{arXiv preprint arXiv:1711.05101}, 2017.

\bibitem[Luo et~al.(2020)Luo, Ji, Shi, Huang, Duan, Li, Li, Bharti, and Zhou]{univl}
Huaishao Luo, Lei Ji, Botian Shi, Haoyang Huang, Nan Duan, Tianrui Li, Jason Li, Taroon Bharti, and Ming Zhou.
\newblock Univl: A unified video and language pre-training model for multimodal understanding and generation.
\newblock \emph{arXiv preprint arXiv:2002.06353}, 2020.

\bibitem[Maaz et~al.(2023)Maaz, Rasheed, Khan, and Khan]{Video-chatgpt}
Muhammad Maaz, Hanoona Rasheed, Salman Khan, and Fahad~Shahbaz Khan.
\newblock Video-chatgpt: Towards detailed video understanding via large vision and language models.
\newblock \emph{arXiv preprint arXiv:2306.05424}, 2023.

\bibitem[Majeedi et~al.(2024)Majeedi, Gajjala, GNVV, and Li]{rica2}
Abrar Majeedi, Viswanatha~Reddy Gajjala, Satya Sai Srinath~Namburi GNVV, and Yin Li.
\newblock Rica\^{} 2: Rubric-informed, calibrated assessment of actions.
\newblock \emph{arXiv preprint arXiv:2408.02138}, 2024.

\bibitem[Matsuyama et~al.(2023)Matsuyama, Kawaguchi, and Lim]{iris}
Hitoshi Matsuyama, Nobuo Kawaguchi, and Brian~Y Lim.
\newblock Iris: Interpretable rubric-informed segmentation for action quality assessment.
\newblock In \emph{Proceedings of the 28th International Conference on Intelligent User Interfaces}, pages 368--378, 2023.

\bibitem[Okamoto and Parmar(2024)]{okamoto2024hierarchical}
Lauren Okamoto and Paritosh Parmar.
\newblock Hierarchical neurosymbolic approach for comprehensive and explainable action quality assessment.
\newblock In \emph{Proceedings of the IEEE/CVF Conference on Computer Vision and Pattern Recognition}, pages 3204--3213, 2024.

\bibitem[Pan et~al.(2019)Pan, Gao, and Zheng]{jrg}
Jia-Hui Pan, Jibin Gao, and Wei-Shi Zheng.
\newblock Action assessment by joint relation graphs.
\newblock In \emph{Proceedings of the IEEE/CVF international conference on computer vision}, pages 6331--6340, 2019.

\bibitem[Papineni et~al.(2002)Papineni, Roukos, Ward, and Zhu]{bleu}
Kishore Papineni, Salim Roukos, Todd Ward, and Wei-Jing Zhu.
\newblock Bleu: a method for automatic evaluation of machine translation.
\newblock In \emph{Proceedings of the 40th annual meeting of the Association for Computational Linguistics}, pages 311--318, 2002.

\bibitem[Parmar and Morris(2019{\natexlab{a}})]{aqa7}
Paritosh Parmar and Brendan Morris.
\newblock Action quality assessment across multiple actions.
\newblock In \emph{2019 IEEE winter conference on applications of computer vision (WACV)}, pages 1468--1476. IEEE, 2019{\natexlab{a}}.

\bibitem[Parmar and Morris(2019{\natexlab{b}})]{mtl-aqa}
Paritosh Parmar and Brendan~Tran Morris.
\newblock What and how well you performed? a multitask learning approach to action quality assessment.
\newblock In \emph{Proceedings of the IEEE/CVF Conference on Computer Vision and Pattern Recognition}, pages 304--313, 2019{\natexlab{b}}.

\bibitem[Parmar and Tran~Morris(2017)]{parmar2017learning}
Paritosh Parmar and Brendan Tran~Morris.
\newblock Learning to score olympic events.
\newblock In \emph{Proceedings of the IEEE conference on computer vision and pattern recognition workshops}, pages 20--28, 2017.

\bibitem[Parmar et~al.(2021)Parmar, Reddy, and Morris]{parmar2021piano}
Paritosh Parmar, Jaiden Reddy, and Brendan Morris.
\newblock Piano skills assessment.
\newblock In \emph{2021 IEEE 23rd international workshop on multimedia signal processing (MMSP)}, pages 1--5. IEEE, 2021.

\bibitem[Parmar et~al.(2022)Parmar, Gharat, and Rhodin]{fitness-aqa}
Paritosh Parmar, Amol Gharat, and Helge Rhodin.
\newblock Domain knowledge-informed self-supervised representations for workout form assessment.
\newblock In \emph{European Conference on Computer Vision}, pages 105--123. Springer, 2022.

\bibitem[Pirsiavash et~al.(2014)Pirsiavash, Vondrick, and Torralba]{mit-skate}
Hamed Pirsiavash, Carl Vondrick, and Antonio Torralba.
\newblock Assessing the quality of actions.
\newblock In \emph{Computer Vision--ECCV 2014: 13th European Conference, Zurich, Switzerland, September 6-12, 2014, Proceedings, Part VI 13}, pages 556--571. Springer, 2014.

\bibitem[Seo et~al.(2022)Seo, Nagrani, Arnab, and Schmid]{seo2022end}
Paul~Hongsuck Seo, Arsha Nagrani, Anurag Arnab, and Cordelia Schmid.
\newblock End-to-end generative pretraining for multimodal video captioning.
\newblock In \emph{Proceedings of the IEEE/CVF Conference on Computer Vision and Pattern Recognition}, pages 17959--17968, 2022.

\bibitem[Shi et~al.(2020)Shi, Ji, Niu, Duan, Zhou, and Chen]{shi2020learning}
Botian Shi, Lei Ji, Zhendong Niu, Nan Duan, Ming Zhou, and Xilin Chen.
\newblock Learning semantic concepts and temporal alignment for narrated video procedural captioning.
\newblock In \emph{Proceedings of the 28th ACM international conference on multimedia}, pages 4355--4363, 2020.

\bibitem[Sun et~al.(2019)Sun, Myers, Vondrick, Murphy, and Schmid]{videobert}
Chen Sun, Austin Myers, Carl Vondrick, Kevin Murphy, and Cordelia Schmid.
\newblock Videobert: A joint model for video and language representation learning.
\newblock In \emph{Proceedings of the IEEE/CVF international conference on computer vision}, pages 7464--7473, 2019.

\bibitem[Tang et~al.(2020)Tang, Ni, Zhou, Zhang, Lu, Wu, and Zhou]{usdl}
Yansong Tang, Zanlin Ni, Jiahuan Zhou, Danyang Zhang, Jiwen Lu, Ying Wu, and Jie Zhou.
\newblock Uncertainty-aware score distribution learning for action quality assessment.
\newblock In \emph{Proceedings of the IEEE/CVF conference on computer vision and pattern recognition}, pages 9839--9848, 2020.

\bibitem[Vaswani et~al.(2017)Vaswani, Shazeer, Parmar, Uszkoreit, Jones, Gomez, Kaiser, and Polosukhin]{transformer}
Ashish Vaswani, Noam Shazeer, Niki Parmar, Jakob Uszkoreit, Llion Jones, Aidan~N Gomez, {\L}ukasz Kaiser, and Illia Polosukhin.
\newblock Attention is all you need.
\newblock \emph{Advances in neural information processing systems}, 30, 2017.

\bibitem[Vedantam et~al.(2015)Vedantam, Lawrence~Zitnick, and Parikh]{cider}
Ramakrishna Vedantam, C Lawrence~Zitnick, and Devi Parikh.
\newblock Cider: Consensus-based image description evaluation.
\newblock In \emph{Proceedings of the IEEE conference on computer vision and pattern recognition}, pages 4566--4575, 2015.

\bibitem[Wang et~al.(2021{\natexlab{a}})Wang, Yang, Zhai, Yu, Suo, Sun, Li, and Zhang]{wang2021survey}
Shunli Wang, Dingkang Yang, Peng Zhai, Qing Yu, Tao Suo, Zhan Sun, Ka Li, and Lihua Zhang.
\newblock A survey of video-based action quality assessment.
\newblock In \emph{2021 International conference on networking systems of AI (INSAI)}, pages 1--9. IEEE, 2021{\natexlab{a}}.

\bibitem[Wang et~al.(2024{\natexlab{a}})Wang, Wang, Yang, Li, Kuang, Zhao, Su, Zhai, and Zhang]{cpr-coach}
Shunli Wang, Shuaibing Wang, Dingkang Yang, Mingcheng Li, Haopeng Kuang, Xiao Zhao, Liuzhen Su, Peng Zhai, and Lihua Zhang.
\newblock Cpr-coach: Recognizing composite error actions based on single-class training.
\newblock In \emph{Proceedings of the IEEE/CVF Conference on Computer Vision and Pattern Recognition}, pages 18782--18792, 2024{\natexlab{a}}.

\bibitem[Wang et~al.(2021{\natexlab{b}})Wang, Zhang, Lu, Zheng, Cheng, and Luo]{pdvc}
Teng Wang, Ruimao Zhang, Zhichao Lu, Feng Zheng, Ran Cheng, and Ping Luo.
\newblock End-to-end dense video captioning with parallel decoding.
\newblock In \emph{Proceedings of the IEEE/CVF international conference on computer vision}, pages 6847--6857, 2021{\natexlab{b}}.

\bibitem[Wang et~al.(2024{\natexlab{b}})Wang, Li, Li, Yu, He, Chen, Pei, Zheng, Xu, Wang, et~al.]{internvideo2}
Yi Wang, Kunchang Li, Xinhao Li, Jiashuo Yu, Yinan He, Guo Chen, Baoqi Pei, Rongkun Zheng, Jilan Xu, Zun Wang, et~al.
\newblock Internvideo2: Scaling video foundation models for multimodal video understanding.
\newblock \emph{arXiv preprint arXiv:2403.15377}, 2024{\natexlab{b}}.

\bibitem[Xia et~al.(2023)Xia, Zhuge, Geng, Fan, Wei, He, and Zheng]{xia2023skating}
Jingfei Xia, Mingchen Zhuge, Tiantian Geng, Shun Fan, Yuantai Wei, Zhenyu He, and Feng Zheng.
\newblock Skating-mixer: Long-term sport audio-visual modeling with mlps.
\newblock In \emph{Proceedings of the AAAI Conference on Artificial Intelligence}, pages 2901--2909, 2023.

\bibitem[Xu et~al.(2022{\natexlab{a}})Xu, Zeng, and Zheng]{lgdt}
Angchi Xu, Ling-An Zeng, and Wei-Shi Zheng.
\newblock Likert scoring with grade decoupling for long-term action assessment.
\newblock In \emph{Proceedings of the IEEE/CVF Conference on Computer Vision and Pattern Recognition}, pages 3232--3241, 2022{\natexlab{a}}.

\bibitem[Xu et~al.(2025)Xu, Ke, Li, Xu, Wu, Lin, and Guo]{xu2025vision}
Huangbiao Xu, Xiao Ke, Yuezhou Li, Rui Xu, Huanqi Wu, Xiaofeng Lin, and Wenzhong Guo.
\newblock Vision-language action knowledge learning for semantic-aware action quality assessment.
\newblock In \emph{European Conference on Computer Vision}, pages 423--440. Springer, 2025.

\bibitem[Xu et~al.(2022{\natexlab{b}})Xu, Rao, Yu, Chen, Zhou, and Lu]{finediving}
Jinglin Xu, Yongming Rao, Xumin Yu, Guangyi Chen, Jie Zhou, and Jiwen Lu.
\newblock Finediving: A fine-grained dataset for procedure-aware action quality assessment.
\newblock In \emph{Proceedings of the IEEE/CVF Conference on Computer Vision and Pattern Recognition}, pages 2949--2958, 2022{\natexlab{b}}.

\bibitem[Xu et~al.(2024{\natexlab{a}})Xu, Huang, Hou, Chen, Zhang, Feng, and Xie]{xu2024retrieval}
Jilan Xu, Yifei Huang, Junlin Hou, Guo Chen, Yuejie Zhang, Rui Feng, and Weidi Xie.
\newblock Retrieval-augmented egocentric video captioning.
\newblock In \emph{Proceedings of the IEEE/CVF Conference on Computer Vision and Pattern Recognition}, pages 13525--13536, 2024{\natexlab{a}}.

\bibitem[Xu et~al.(2024{\natexlab{b}})Xu, Yin, Zhao, Wang, and Peng]{fineparser}
Jinglin Xu, Sibo Yin, Guohao Zhao, Zishuo Wang, and Yuxin Peng.
\newblock Fineparser: A fine-grained spatio-temporal action parser for human-centric action quality assessment.
\newblock In \emph{Proceedings of the IEEE/CVF Conference on Computer Vision and Pattern Recognition}, pages 14628--14637, 2024{\natexlab{b}}.

\bibitem[Yamazaki et~al.(2023)Yamazaki, Vo, Truong, Raj, and Le]{yamazaki2023vltint}
Kashu Yamazaki, Khoa Vo, Quang~Sang Truong, Bhiksha Raj, and Ngan Le.
\newblock Vltint: Visual-linguistic transformer-in-transformer for coherent video paragraph captioning.
\newblock In \emph{Proceedings of the AAAI Conference on Artificial intelligence}, pages 3081--3090, 2023.

\bibitem[Yang et~al.(2023)Yang, Nagrani, Seo, Miech, Pont-Tuset, Laptev, Sivic, and Schmid]{vid2seq}
Antoine Yang, Arsha Nagrani, Paul~Hongsuck Seo, Antoine Miech, Jordi Pont-Tuset, Ivan Laptev, Josef Sivic, and Cordelia Schmid.
\newblock Vid2seq: Large-scale pretraining of a visual language model for dense video captioning.
\newblock In \emph{Proceedings of the IEEE/CVF Conference on Computer Vision and Pattern Recognition}, pages 10714--10726, 2023.

\bibitem[Yin et~al.(2025)Yin, Parmar, Xu, Zhang, Zheng, and Fu]{yin2025decade}
Hao Yin, Paritosh Parmar, Daoliang Xu, Yang Zhang, Tianyou Zheng, and Weiwei Fu.
\newblock A decade of action quality assessment: Largest systematic survey of trends, challenges, and future directions.
\newblock \emph{arXiv preprint arXiv:2502.02817}, 2025.

\bibitem[Yu et~al.(2021)Yu, Rao, Zhao, Lu, and Zhou]{core}
Xumin Yu, Yongming Rao, Wenliang Zhao, Jiwen Lu, and Jie Zhou.
\newblock Group-aware contrastive regression for action quality assessment.
\newblock In \emph{Proceedings of the IEEE/CVF international conference on computer vision}, pages 7919--7928, 2021.

\bibitem[Yun et~al.(2024)Yun, Qi, Peng, and Ma]{yun2024semi}
Wulian Yun, Mengshi Qi, Fei Peng, and Huadong Ma.
\newblock Semi-supervised teacher-reference-student architecture for action quality assessment.
\newblock In \emph{European Conference on Computer Vision}, pages 161--178. Springer, 2024.

\bibitem[Zeng and Zheng(2024)]{zeng2024multimodal}
Ling-An Zeng and Wei-Shi Zheng.
\newblock Multimodal action quality assessment.
\newblock \emph{IEEE Transactions on Image Processing}, 2024.

\bibitem[Zeng et~al.(2020)Zeng, Hong, Zheng, Yu, Zeng, Wang, and Lai]{zeng2020hybrid}
Ling-An Zeng, Fa-Ting Hong, Wei-Shi Zheng, Qi-Zhi Yu, Wei Zeng, Yao-Wei Wang, and Jian-Huang Lai.
\newblock Hybrid dynamic-static context-aware attention network for action assessment in long videos.
\newblock In \emph{Proceedings of the 28th ACM international conference on multimedia}, pages 2526--2534, 2020.

\bibitem[Zhang et~al.(2023{\natexlab{a}})Zhang, Dai, Wang, Shen, Lu, Zhou, and Tang]{logo}
Shiyi Zhang, Wenxun Dai, Sujia Wang, Xiangwei Shen, Jiwen Lu, Jie Zhou, and Yansong Tang.
\newblock Logo: A long-form video dataset for group action quality assessment.
\newblock In \emph{Proceedings of the IEEE/CVF Conference on Computer Vision and Pattern Recognition}, pages 2405--2414, 2023{\natexlab{a}}.

\bibitem[Zhang et~al.(2024)Zhang, Bai, Chen, Chen, Lu, Wang, and Tang]{nae}
Shiyi Zhang, Sule Bai, Guangyi Chen, Lei Chen, Jiwen Lu, Junle Wang, and Yansong Tang.
\newblock Narrative action evaluation with prompt-guided multimodal interaction.
\newblock In \emph{Proceedings of the IEEE/CVF Conference on Computer Vision and Pattern Recognition}, pages 18430--18439, 2024.

\bibitem[Zhang et~al.(2023{\natexlab{b}})Zhang, Pan, Gao, and Zheng]{shaojie-tmm}
Shao-Jie Zhang, Jia-Hui Pan, Jibin Gao, and Wei-Shi Zheng.
\newblock Adaptive stage-aware assessment skill transfer for skill determination.
\newblock \emph{IEEE Transactions on Multimedia}, 2023{\natexlab{b}}.

\bibitem[Zhang et~al.(2019)Zhang, Kishore, Wu, Weinberger, and Artzi]{bertscore}
Tianyi Zhang, Varsha Kishore, Felix Wu, Kilian~Q Weinberger, and Yoav Artzi.
\newblock Bertscore: Evaluating text generation with bert.
\newblock \emph{arXiv preprint arXiv:1904.09675}, 2019.

\bibitem[Zhou et~al.(2024{\natexlab{a}})Zhou, Cai, Wang, Shum, and Liang]{zhou2024comprehensive}
Kanglei Zhou, Ruizhi Cai, Liyuan Wang, Hubert~PH Shum, and Xiaohui Liang.
\newblock A comprehensive survey of action quality assessment: Method and benchmark.
\newblock \emph{arXiv preprint arXiv:2412.11149}, 2024{\natexlab{a}}.

\bibitem[Zhou et~al.(2024{\natexlab{b}})Zhou, Wang, Zhang, Shum, Li, Li, and Liang]{zhou2024magr}
Kanglei Zhou, Liyuan Wang, Xingxing Zhang, Hubert~PH Shum, Frederick~WB Li, Jianguo Li, and Xiaohui Liang.
\newblock Magr: Manifold-aligned graph regularization for continual action quality assessment.
\newblock In \emph{European Conference on Computer Vision}, pages 375--392. Springer, 2024{\natexlab{b}}.

\end{thebibliography}
}

\newpage
\appendix
\section{Appendix}
In this appendix, we provide more details about the evaluation metrics, implementations, experiment results and analysis in \cref{sec:appendix_exp}. Subsequently, more details about the EE4D-DescCoach dataset are introduced in \cref{sec:appendix_dataset}, including examples of general technical points (TechPoints), hierarchical coaching commentary, prompts used for LLM-based annotation collection, and quality ensurance.
Finally, we discuss the limitations and potential directions for future work in \cref{sec:appendix_dicussion}.

\section{More Details about Experiments}
\label{sec:appendix_exp}
\subsection{More Details about the Evaluation Metrics}
\noindent{\textbf{Spearman's rank correlation coefficient}} ($\rho$) \cite{wang2021survey} is the most widely used metric to evaluate models in the field of AQA. It measures the correlation between the predicted score series and the ground-truth score series, which is defined as follows:
\begin{equation}
    \rho= \frac{\sum^N_{i=1}(x_i-\bar x)(y_i - \bar y)}{\sqrt{\sum^N_{i=1}(x_i-\bar x)^2\sum^N_{i=1}(y_i - \bar y)^2}},
\end{equation}
where $x$ and $y$ are the rankings of the two series.

\vspace{0.1cm}
\noindent{\textbf{Relative-L2 Distance}} (RL2) \cite{core} evaluates the numerical differences between the predicted and ground-truth scores, which can be written as:
\begin{equation}
    RL2 = \frac{1}{N} \sum^N_{n=1}(\frac{|s_n - \hat{s}_n|}{s_{max}-s_{min}})^2,
\end{equation}
where $s_n$ and $\hat s_n$ indicate the ground-truth and predicted scores, respectively. 
$s_{max}$ and $s_{min}$ indicate the highest and lowest scores for an action instance.

\vspace{0.1cm}
\noindent{\textbf{BLEU}} \cite{bleu} quantifies the overlap between the generated sentences and the reference sentences by comparing n-grams (contiguous word sequences) shared between them. In this work, we compute the 1-gram-based BLEU score to evaluate the models.

\vspace{0.1cm}
\noindent{\textbf{METEOR}} \cite{meteor} improves upon BLEU by introducing flexible word matching (e.g., synonyms and stemming) to better capture meaning beyond exact matches. Additionally, it adopts a recall-oriented approach to ensure all reference words are considered, and incorporates word order penalties to evaluate fluency and coherence.

\vspace{0.1cm}
\noindent{\textbf{CIDEr}} \cite{cider} is originally proposed for image captioning task. It measures the similarity between generated and reference texts using n-gram overlap weighted by TF-IDF. This ensures important but less common phrases are emphasized, rewarding consensus among references while penalizing overly generic outputs.

\vspace{0.1cm}
\noindent{\textbf{BERTScore}} \cite{bertscore} evaluates natural language generation by comparing the semantic similarity between generated and reference texts using contextual embeddings from pre-trained BERT-like models. Unlike traditional metrics that rely on n-gram overlap, BERTScore computes token-level similarity based on meaning, capturing nuances like synonyms and paraphrasing. In our work, we use the DeBERTa-xlarge \cite{deberta} to extract token-level embeddings as it shows great correlation with human evaluation.

\begin{figure*}[t]
    \centering
    \includegraphics[width=1\linewidth]{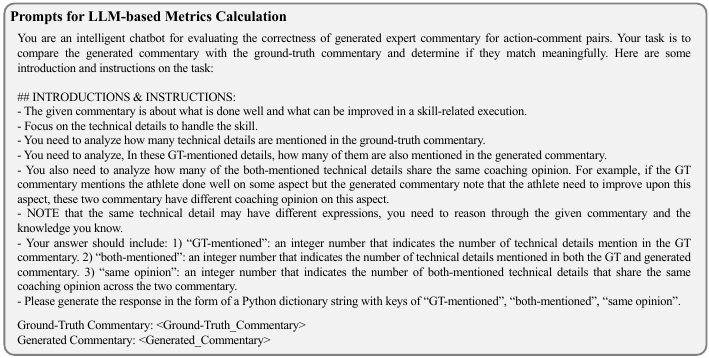}
    \vspace{-0.7cm}
    \caption{\textbf{The detailed prompts for LLM-based evaluation}. Zoom in for best view.}
    \label{fig:prompts_llm_eval}
    \vspace{-0.4cm}
\end{figure*}

\begin{figure}[t]
    \centering
    \includegraphics[width=1\linewidth]{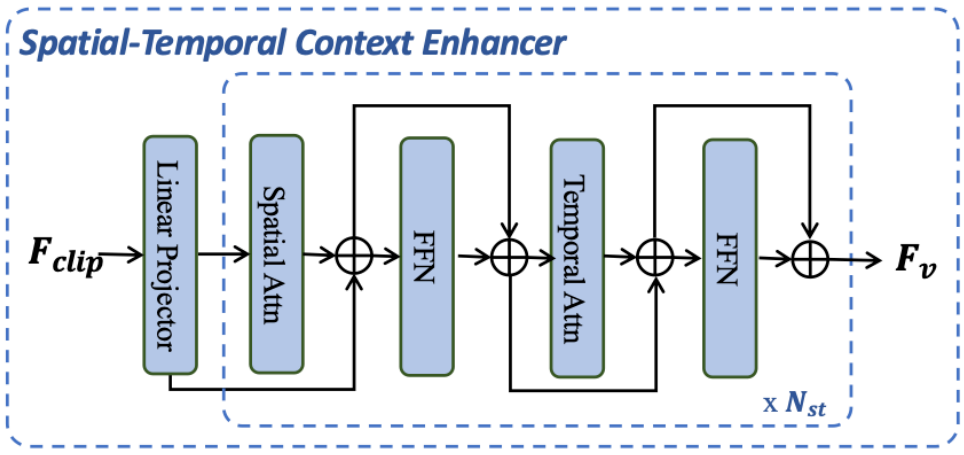}
    \vspace{-0.7cm}
    \caption{\textbf{An overview of the Spatial-Temporal Context Enhancer}. Here the $\oplus$ indicates the element-level adding operation.}
    \label{fig:st_enhancer}
    \vspace{-0.4cm}
\end{figure}

\vspace{0.1cm}
\noindent{\textbf{GPT-M}} and  \textbf{GPT-Q} are the LLM-based metrics proposed to evaluate whether the generated commentary shares the same technical details and the same coaching opinions as the ground-truth commentary. We employ the GPT-4o-mini to determine the number of both-mentioned technical details and the shared coaching opinions. Detailed prompts are provided in \cref{fig:prompts_llm_eval}. After that, the GPT-M score is calculated as the averaged proportion of the both-mentioned technical details to those mentioned by ground truth across all testing instances. The GPT-Q score is calculated as the averaged proportion of the technical details sharing the same coaching opinions to the both-mentioned technical details.

\begin{figure*}[t]
    \vspace{-0.3cm}
    \centering
    \includegraphics[width=0.7\linewidth]{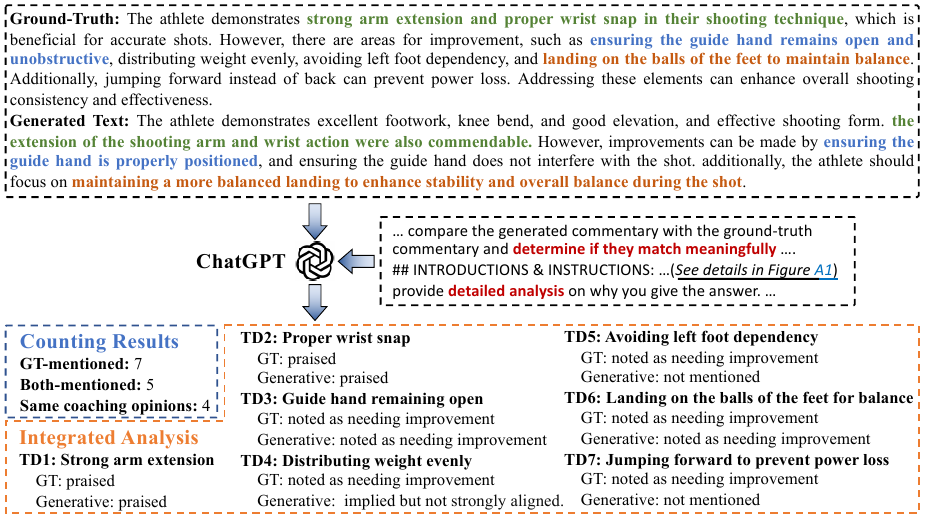}
    \vspace{-0.3cm}
    \caption{\textbf{Qualitative results on the LLM-based evaluation}. We take genertaedt and ground-truth commentary from Fig. \ym{5} in the main paper, and prompt LLM to produce accurate counts along with detailed analysis. TD: recognized technical details.}
    \label{fig:gpt_eval_visualization}
    \vspace{-0.3cm}
\end{figure*}

\subsection{More Implementation Details}
\noindent{\textbf{- Spatial-Temporal Context Enhancer}}. 
As discussed in Sec. \ym{4.3} of the main paper, inspired by TimeSformer \cite{timesformer}, we adopt a Spatial-Temporal Context Enhancer to enhance the spatial-temporal context across the clip-level features. An overview of this module is shown in \cref{fig:st_enhancer}. Specifically, we use a linear projector to map the clip-level features $F_{clip}$ to the hidden dimension $D$, and feed them into $N_{st}$ Transformer blocks that include spatial- and temporal-attention layers along with the FFN layers. In our implementation, the $N_{st}$ is set to 4.

\vspace{0.1cm}
\noindent{\textbf{- Other details of TechCaoch}}. 
The visual backbone and text encoder are initialized by the pretrained weights and frozen during training. For the visual backbone, we use the pretrained weights of InternVideo2-stage3 \cite{internvideo2}. We use the pretrained DeBERTa-xlarge \cite{deberta} as the text encoder.
Following \cite{swinbert, nae}, we employ AdamW \cite{adamw} as the optimizer, and utilize a learning rate warm-up during the early 10\% training steps followed by linear decay. 
$\lambda_1$ and $\lambda_2$ are set to 0.1 and 0.3, respectively.
Due to the characteristics of EgoExo4D \cite{egoexo4d} and the annotation pipeline of EE4D-DescCoach, it is not possible to ensure each instance includes rounded commentary on every possible TechPoints (\eg, in some cases the coaching commentary would not mention the execution about the \emph{head \& eyes}). 
Therefore, we make some adjustments on the prompts or instructions for the compared models to guide the models to focus on those TechPoints mentioned in the ground-truth Commentary.
For TechCoach, we employ zero-masking on the TechPoint-related quality embeddings corresponded to a specific TechPoint only if both the strength and weakness on this TechPoint are unmentioned in the ground truth. Note that such a design does NOT leak the action quality information, and similar adjustments are applied on the compared models \cite{nae,video-llava,videochat2,internvideo2,internvl2} for fair comparisons. 

\vspace{0.1cm}
\noindent{\textbf{- Details about USDL \cite{usdl}}}. USDL proposes to convert the score regression problem to a distribution prediction problem. We primarily follow the official implementation. We make adjustments on the output dimension to align with the score range in our experiments.

\vspace{0.1cm}
\noindent{\textbf{- Details about CoRe \cite{core} and TPT \cite{tpt}}}. These two methods adopt the Contrastive Regression paradigm to generate action quality scores by predicting the score differences between the query instance and multiple exemplar instances. Our experiments are conducted based on the official implementations of these methods. The exemplars are randomly selected from the training instances with the same action task (\eg, Reverse Layup) as the query instance. The number of selected exemplars is set to 10.

\vspace{0.1cm}
\noindent{\textbf{- Details about SwinBERT \cite{swinbert}}}. SwinBERT proposes a general framework for video captioning. In addition to the modification of the visual backbone, we also modified the length of masking tokens during training to ensure consistency with our TechCoach. Additionally, following \cite{nae}, we add an extra full-connected layer to regress the score.

\vspace{0.1cm}
\noindent{\textbf{- Details about PGMI \cite{nae}}}. PGMI is a prompt-guided framework for Multi-Task Learning AQA paradigm. Our experiments on PGMI are conducted using the official implementation. We adjust the number of score groups to 10 as the score range in our experiments is in [0, 10]. Besides, we modify the input templates for the text generator to include the score information and the technical dimensions that should be focused on.
Specifically, the template is adjusted as follows: 

``\emph{The player is practicing \textless Action\_Task\textgreater, and gets a score of \textless Predicted\_Score\textgreater. Pay attention to the execution of \textless Technical\_Dimensions\textgreater}'', 

\noindent where \emph{\textless Action\_Task\textgreater} is the action task, \emph{\textless Predicted\_Score\textgreater} indicates the predicted score by the model, and \emph{\textless Technical\_Dimensions\textgreater} indicates the a list of technical dimensions (\eg, \emph{head \& eyes}, \emph{arms \& elbows}, and \emph{human-object interaction}) whose corresponded TechPoints are mentioned in the ground truth.

\vspace{0.1cm}
\noindent{\textbf{- Details about the Compared MLLMs \cite{video-llava,videochat2,internvideo2,internvl2}}}. We follow the official guidelines to perform zero-shot evaluation on the compared MLLMs. 
A general language instruction for these methods is as follows: 

``\emph{Suppose you are an \textless Action\_Task\textgreater  coach. Watch the given video, and answer a paragraph of commentary about what is done well in the execution and what can be improved. Your answer should be like: \textless Answer\_Example\textgreater. Pay attention to the execution of \textless Technical\_Dimensions\textgreater }'', 

\noindent where \emph{\textless Action\_Task\textgreater} is the action task (\eg, Reverse Layup), \emph{\textless Answer\_Example\textgreater} is an example of the answer (selected from training set), and \emph{\textless Technical\_Dimensions\textgreater} indicates the a list of technical dimensions (\eg, \emph{head \& eyes}, \emph{arms \& elbows}, and \emph{human-object interaction}) whose corresponded TechPoints are mentioned in the ground-truth commentary.

In \cref{sec:more_quantitative_results}, we also explore building a DescCoach-specific MLLM. Specifically, we take VideoChat2 \cite{videochat2} as an example, and fine-tune it with LoRA \cite{lora} while keep the other parameters frozen on our EE4D-DescCoach dataset.

\begin{figure}[t]
    \vspace{-0.3cm}
    \centering
    \includegraphics[width=1\linewidth]{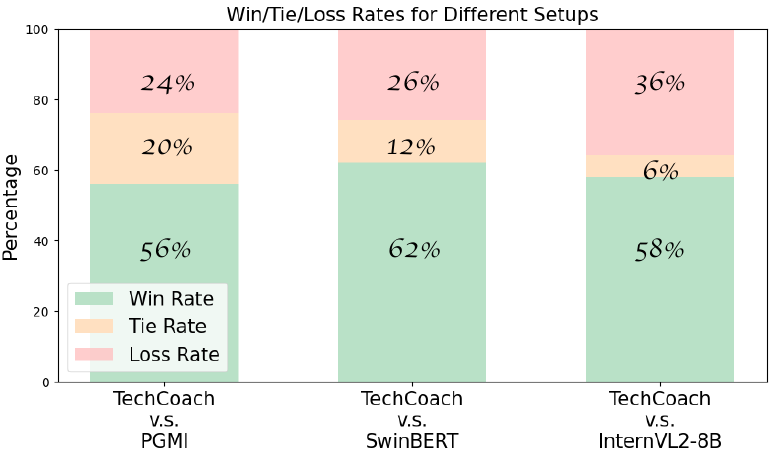}
    \vspace{-0.6cm}
    \caption{\textbf{User study}. Our TechCoach wins the compared baselines frequently in human evaluation.}
    \label{fig:user_study}
    \vspace{-0.3cm}
\end{figure}

\begin{figure}[t]
    \vspace{-0.3cm}
    \centering
    \includegraphics[width=1\linewidth]{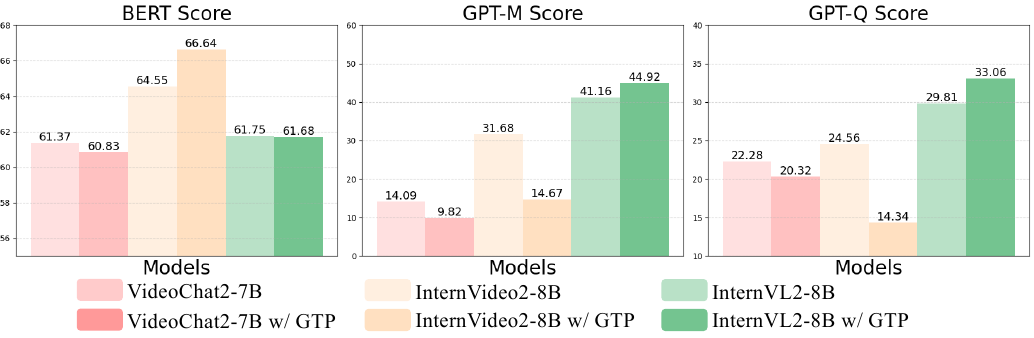}
    \vspace{-0.6cm}
    \caption{\textbf{Qualitative results on the LLM-based evaluation}. We take genertaedt and ground-truth commentary from Fig. \ym{5} in the main paper, and prompt LLM to produce accurate counts along with detailed analysis}
    \label{fig:gtp_in_prompt}
    \vspace{-0.3cm}
\end{figure}

\begin{table}[t]
    \scriptsize
    \centering
    \scriptsize
    \resizebox{1\linewidth}{!}{
        \begin{NiceTabular}{l|cccccc}
        \toprule
        \multirow{2}{*}{\textbf{{Methods}}} &  \multicolumn{6}{c}{\textbf
{Commentary}}   \\ 
         & B$\uparrow$ & C$\uparrow$ & M$\uparrow$ & Bert$\uparrow$ & GPT-M$\uparrow$ &
GPT-Q$\uparrow$\\
        \midrule
        VideoChat2-7B \cite{videochat2}&  23.18 & 2.98 & 11.59 & 61.37 & 14.09 & 22.28\\
        
        VideoChat2-7B-FT \cite{videochat2}&  28.74 & 4.88 & 13.27 & 62.98 & 36.32 & 23.62\\
        \midrule
        \rowcolor{liym_yellow}  TechCoach(Ours) &  \textbf{37.06} & \textbf{14.62} & \textbf{16.39}  & \textbf{66.89} & \textbf{50.44} & \textbf{38.15}\\
        \bottomrule        
        \end{NiceTabular}
    }
    \vspace{-0.3cm}
    \caption{\textbf{Comparison with fine-tuned MLLM.} While finetuning VideoChat2 on our EE4D-DescCoach dataset bring stable improvement on all metrics, the modest gain on GPT-Q score indicates that the model still struggle to obtain strong action quality perceiving ability. FT: finetuned on our EE4D-DescCoach dataset.}
    \label{tab:ft_llm}
\end{table}

\begin{table}[t]
    \scriptsize
    \centering
    \scriptsize
    \resizebox{1\linewidth}{!}{
        \begin{NiceTabular}{l|cc|cccc}
        \toprule
        \multirow{2}{*}{\textbf{Variants}} & \multicolumn{2}{c|}{\textbf{Score}}  & \multicolumn{4}{c}{\textbf{Commentary}}  \\ 
        & $\rho\uparrow$ & RL2$\downarrow$ & B$\uparrow$ & C$\uparrow$ & M$\uparrow$
         & BERT$\uparrow$ \\
        \midrule
        InfoNCE-Align & 71.38 & 4.60 & 36.76 & 13.72 & 16.26 & 65.78\\
        MarginRanking-Align & 71.67 & 4.36 & 36.96 & 14.29 & 16.35 & 66.45 \\
        \rowcolor{liym_yellow} L2-Align & \textbf{72.46} & \textbf{4.35}& \textbf{37.06} & \textbf{14.62} & \textbf{16.39} & \textbf{66.89}  \\
        \bottomrule        
        \end{NiceTabular}
    }
    \vspace{-0.3cm}
    \caption{\textbf{Comparison with variants with other types of feature alignment losses}. TechCoach achieves
    stronger performance with simple yet effective L2-distance-based alignment loss.}
    \vspace{-0.3cm}
    \label{tab:ablation_alignment_loss}
\end{table}

\begin{table}[t]
    \scriptsize
    \centering
    \scriptsize
    \resizebox{1\linewidth}{!}{
        \begin{NiceTabular}{l|cc|cccc}
        \toprule
        \multirow{2}{*}{\textbf{Variants}} & \multicolumn{2}{c|}{\textbf{Score}}  & \multicolumn{4}{c}{\textbf{Commentary}}  \\ 
        & $\rho\uparrow$ & RL2$\downarrow$ & B$\uparrow$ & C$\uparrow$ & M$\uparrow$
         & Bert$\uparrow$ \\
        \midrule
        Ego-Only  & 69.31 & 4.75 & 36.87 & 13.79 & 16.16 & 66.13\\
        Exo-Only & 68.55 & 4.84 & \textbf{37.31} & 12.71 & 16.20 & 66.23 \\
        \rowcolor{liym_yellow} Ego+Exo & \textbf{72.46} & \textbf{4.35}& {37.06} & \textbf{14.62} & \textbf{16.39} & \textbf{66.89}  \\
        \bottomrule        
        \end{NiceTabular}
    }
    \vspace{-0.3cm}
    \caption{\textbf{Comparison with Ego- and Exo-only variants}. By collaborating on both Ego and Exo videos, TechCoach achieves stronger performance, showcasing the complementary role across different views for descriptive action coaching.}
    \vspace{-0.3cm}
    \label{tab:ablation_views}
\end{table}

\subsection{More Comparison Results and Analysis}
\label{sec:more_quantitative_results}

\noindent{\textbf{- User study}}. 
To further compare the performance across different models, we conducted a comprehensive user study comparing our TechCoach against three baselines (\ie, PGMI\cite{nae}, SwinBERT\cite{swinbert}, and InternVL2-8B\cite{internvl2}). 
We randomly select 50 instances from evaluation set and ask all four models to generate parallel outputs. We recruit five labelers (not involved in our project) and each of them is assigned instances through random sampling to eliminate evaluation bias.
We provide similar assessment instructions as those for LLM evaluation (see \cref{fig:prompts_llm_eval}), and ask the labelers performed blind assessments by ranking the four outputs per instance based on semantic alignment with ground-truth (GT) text, allowing tied rankings (e.g., TechCoach=PGMI{\textgreater}InternVL2-8B{\textgreater}SwinBERT). 

After that, we computed pairwise comparison metrics:
(1) \emph{Win Rate}: Percentage of instances where TechCoach outperforms the baseline;
(2) \emph{Tie Rate}: Percentage of instances where there is a tie between TechCoach and the baseline;
(3) \emph{Loss Rate}: Percentage of instances where the baseline model prevail.
The user study results are shown in \cref{fig:user_study}. Compared with other baselines, our TechCoach achieves clearly higher \emph{Win Rate}, further illustrating the generated text of TechCoach can effectively catch the semantic meaning mentioned in Groun-Truth and align with human preference.

\vspace{0.1cm}
\noindent{\textbf{- Adding general TechPoints into the prompts for MLLMs}}.
As discussed in Sec. \ym{4.2}, our TechCoach inferences with assumption that general TechPoints (GTP) are available. To evaluate the influence of GTP for general MLLMs, we try to add general TechPoints into the prompts for MLLMs. 
As shown in \cref{fig:gtp_in_prompt}, adding the (GTP) into the prompts brings unconsistent performance effects across different MLLMs (\ie, InternVL2-8B gets improvement but VideoChat2-7B and InternVideo2 gets opposite effects). 
These results suggest that a more effective approach to integrating TechPoints into the action assessment process is necessary.

\vspace{0.1cm}
\noindent{\textbf{- Fine-tuning MLLM on EE4D-DescCoach dataset}}. As another branch of exploration, we try to fine-tune MLLM on our EE4D-DescCoach dataset. We take VideoChat2 \cite{videochat2} as an example, and fine-tune it with LoRA \cite{lora}. The results are shown in \cref{tab:ft_llm}. While finetuning VideoChat2 on our EE4D-DescCoach dataset brings stable improvement on all metrics, the modest gain on GPT-Q score indicates that the model still struggle to obtain strong action quality perceiving ability.

\vspace{0.1cm}
\noindent{\textbf{- Why does InternVL2 achieve abnormal performance on traditional metrics?}} In Tab. \ym{2} of main paper, InternVL2 \cite{internvl2} family models obtain abnormal performance on traditional n-gram based metrics (\ie, BLEU, METEOR, CIDEr) and feature similarity-based metric (BERTScore). By going deeper to the outputs of InternVL2, we find the reasons behind it could be: even if an answer sample is provided in the prompt, InternVL2 tends to provide answers in a fixed, list-like format. For example: 

``\emph{\#\# Execution Done Well: 1. {**}Starting Position{**}: The player demonstrates a strong starting position with ... 2. **Ball Handling**: The dribbling technique appears consistent ...}''

The formats of answers and ground-truth commenatry are highly inconsistent, which makes the traditional metrics fail to effectively evaluate the model performance.

\subsection{More Ablation Studies}
\vspace{0.1cm}
\noindent{\textbf{- Impacts of various types of TechPoint-level alignment loss.}} As discussed in Sec. \ym{4.4}, L2 distance is used to measure the distance between TechPoint-related quality embeddings and TechPoint-level commentary features. We further investigate the imapcts of alignment loss by introducing two other kinds of alignment losses, i.e., InfoNCE and MarginRanking. As shown in \cref{{tab:ablation_alignment_loss}}, TechCoach achieves better performance with the L2 distance-based alignment loss. 

\vspace{0.1cm}
\noindent{\textbf{- Impacts of videos from various views.}}
As discussed in Sec. \ym{5.2}, we train TechCoach with both Egocentric and Exocentric videos. Here we conduct an experiment on the influence of the training videos from different views. As shown in \cref{tab:ablation_views}, Ego-Only and Exo-Only variants showcase comparable performance. By combining videos from different views, TechCoach achieves stronger performance, which shows that egocentric and exocentric viewpoints are complementary for action coaching.

\subsection{More Qualitative Results}
\label{sec:more_visualization}
\noindent\textbf{- LLM can provide a reliable evaluation of generated text.} To highlight this, we take an example from Fig. \ym{5} in the main paper, and prompt LLM to produce accurate counts along with detailed analysis, which are shown in \cref{fig:gpt_eval_visualization}. 

\begin{figure*}[t]
    \centering
    \includegraphics[width=1\linewidth]{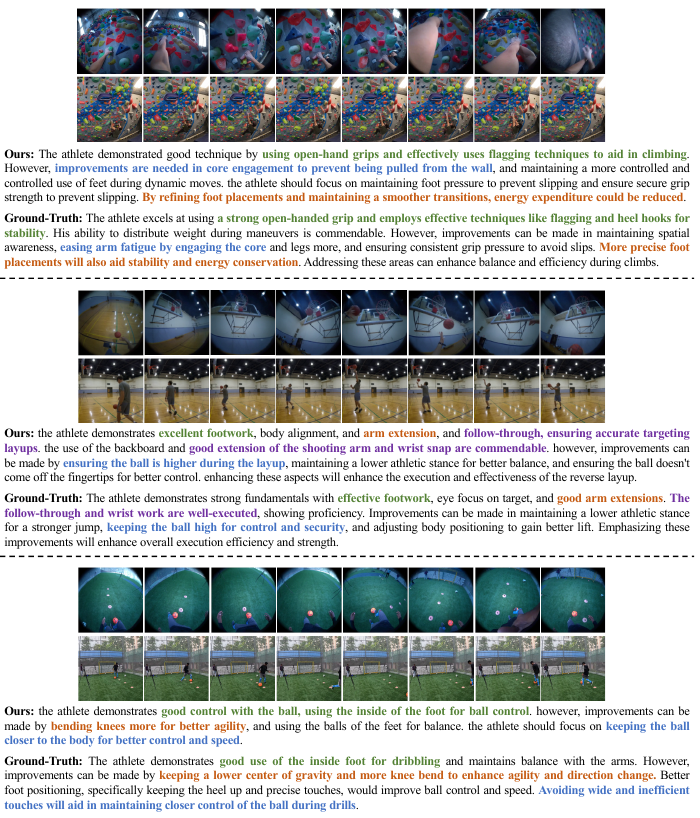}
    \caption{\textbf{More qualitative results on the generated coaching commentary}. Our TechCoach generates precise and detailed coaching commentary on \textbf{\emph{what is done well}} and \textbf{\emph{what can be improved}} from the given action videos. Correctly matched parts are highlighted in colors.}
    \label{fig:qualitative_results_appendix}
\end{figure*}

\begin{figure*}[t]
    \centering
    \includegraphics[width=0.9\linewidth]{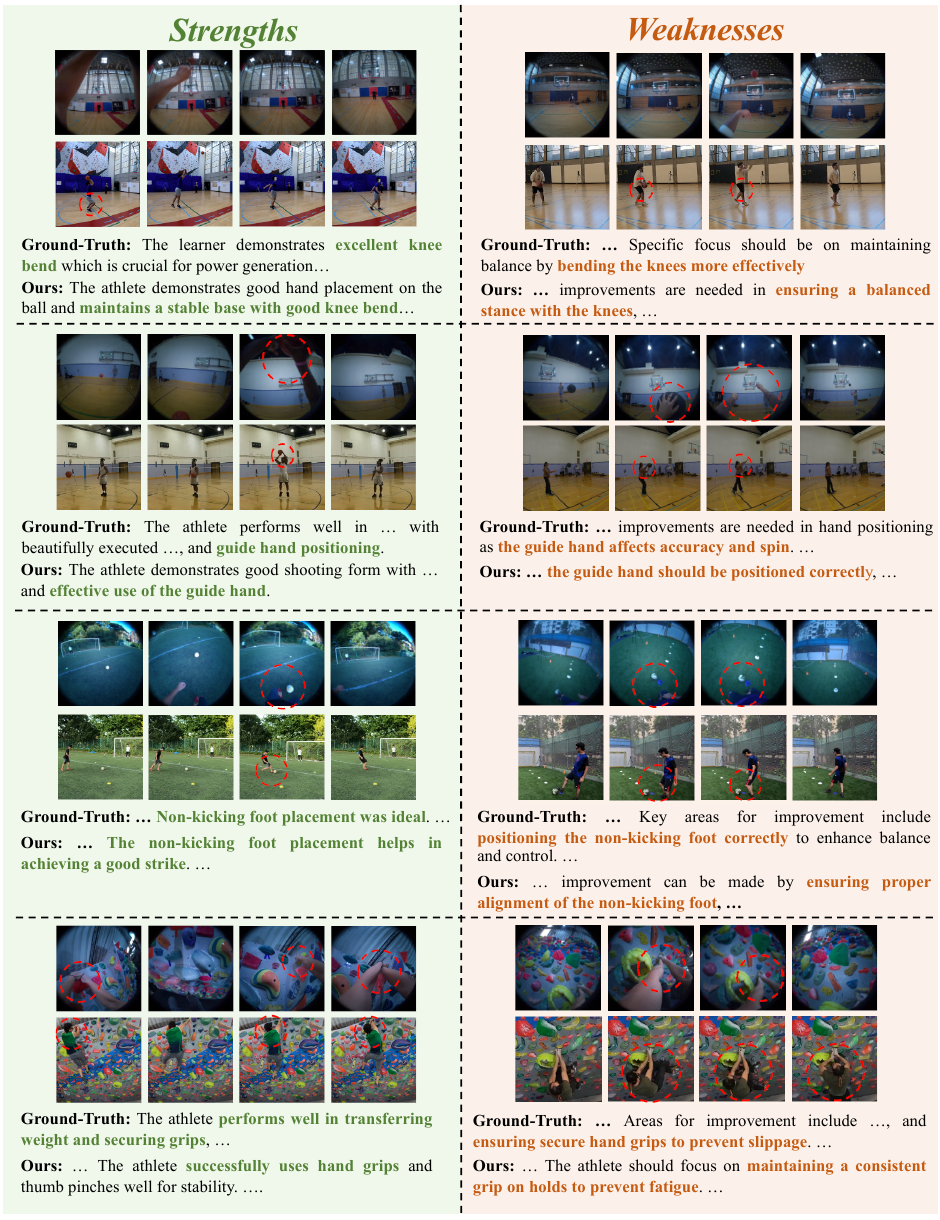}
    \vspace{-0.3cm}
    \caption{\textbf{More qualitative results on the generated coaching commentary.} Our TechCoach is capable of perceiving the action quality (\ie, \textcolor{strength_green}{\textbf{\emph{Strengths}}} and \textcolor{weakness_orange}{\textbf{\emph{Weaknesses}}}) of the same technical details (\eg, bending the knee) across various action executions. Correctly matched parts are highlighted in colors, and the corresponding regions are marked with dashed red circle. Some frames are zoomed in for better visibility.}
    \label{fig:qualitative_results_quality_appendix}
\end{figure*}

\vspace{0.1cm}
\noindent\textbf{- TechCoach is able to perceive the action quality}. As a supplement to Fig. \ym{5} in the main paper, we present additional qualitative results on the generated coaching commentary in \cref{fig:qualitative_results_appendix}, further illustrating the effectiveness of our TechCoach.
Moreover, \cref{fig:qualitative_results_quality_appendix} demonstrates that our TechCoach is capable of perceiving the action quality (\ie, \emph{Strengths} and \emph{Weaknesses}) of the same technical details on various action executions.

\section{More Details about EE4D-DescCoach}
\label{sec:appendix_dataset}
In this section, we show the examples of the obtained general TechPoints, hierarchical coaching commentary, prompts for annotation collection, and other details.

\vspace{0.1cm}
\noindent{\textbf{- General TechPoints}}. Completing a skilled physical action typically requires coordination across multiple body parts and objects. Thus, we define TechPoints as language instructions spanning seven technical dimensions: (1) \emph{Head \& Eyes}; (2) \emph{Torso \& Core}; (3) \emph{Arms \& Elbows}; (4) \emph{Wrists \& Hands}; (5) \emph{Legs \& Knees}; (6) \emph{Feet \& Heels}; and (7) \emph{Human-Object Interaction}. \cref{tab:all_TKP_mikan_layup}, \cref{tab:all_TKP_climbing} and \cref{tab:all_TKP_juggling} provide examples of the general TechPoints for specific action tasks. 

\begin{figure*}[t]
    \centering
    \includegraphics[width=0.86\linewidth]{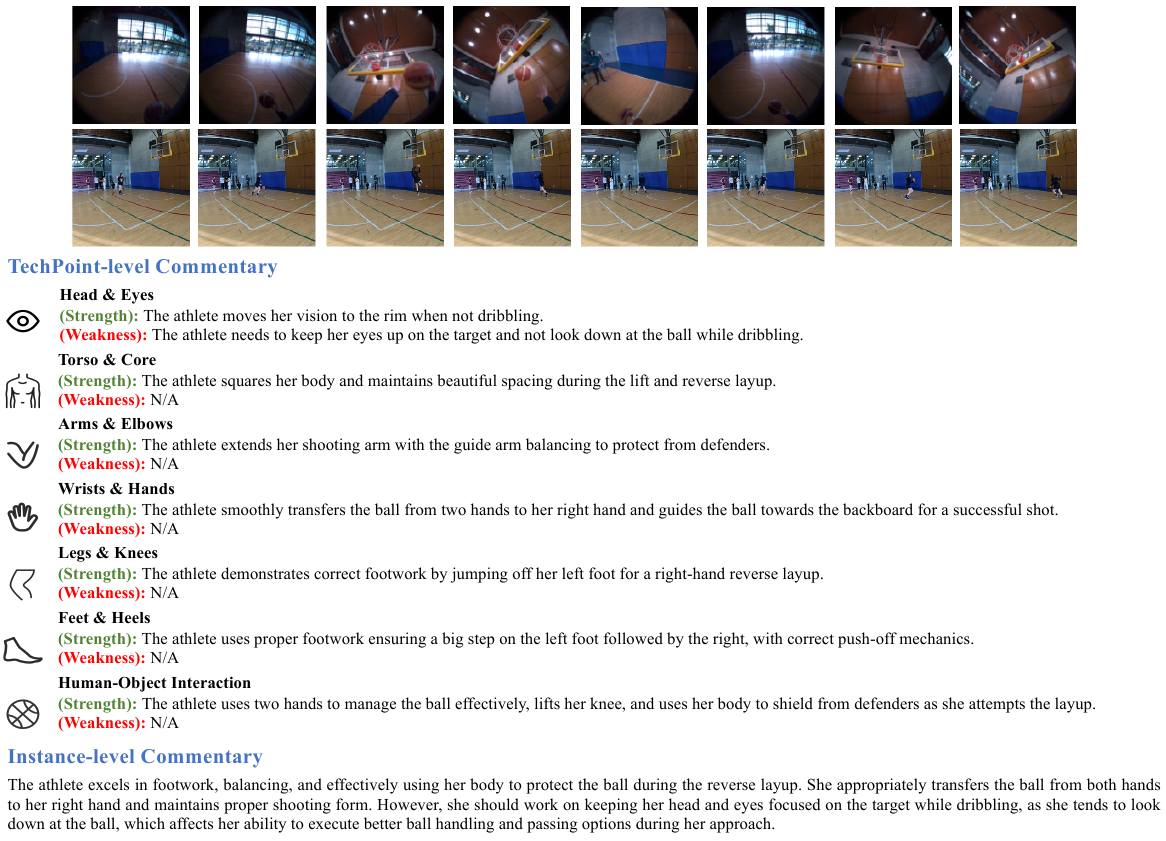}
    \vspace{-0.3cm}
    \caption{\textbf{An example for hierarchical coaching commentary.} Zoom in for best view.}
    \label{fig:hier_com_Basketball}
    \vspace{-0.25cm}
\end{figure*}

\begin{figure*}[t]
    \centering
    \includegraphics[width=0.85\linewidth]{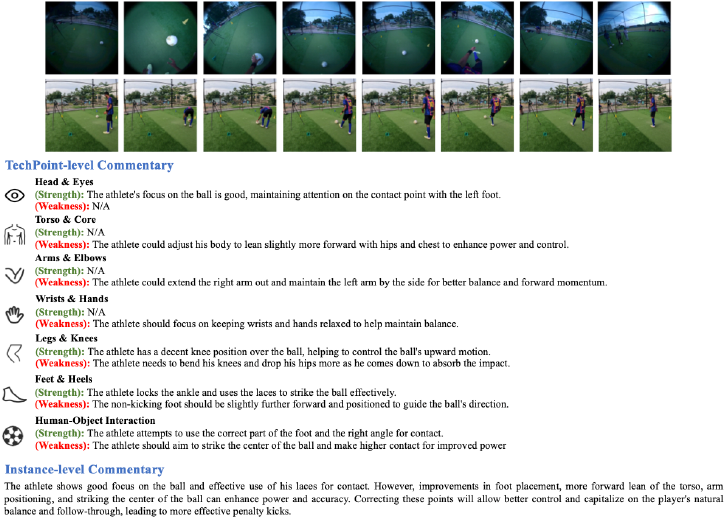}
    \vspace{-0.3cm}
    \caption{\textbf{An example for hierarchical coaching commentary.} Zoom in for best view.}
    \label{fig:hier_com_Soccer}
    \vspace{-0.3cm}
\end{figure*}

\vspace{0.1cm}
\noindent{\textbf{- Hierarchical Coaching Commentary}}. Our EE4D-DescCoach dataset features hierarchical coaching commentary on both TechPoint level and instance level. 
\cref{fig:hier_com_Basketball} and \cref{fig:hier_com_Soccer} show examples of the hierarchical coaching commentary.

\begin{figure*}[t]
    \centering
    \includegraphics[width=1\linewidth]{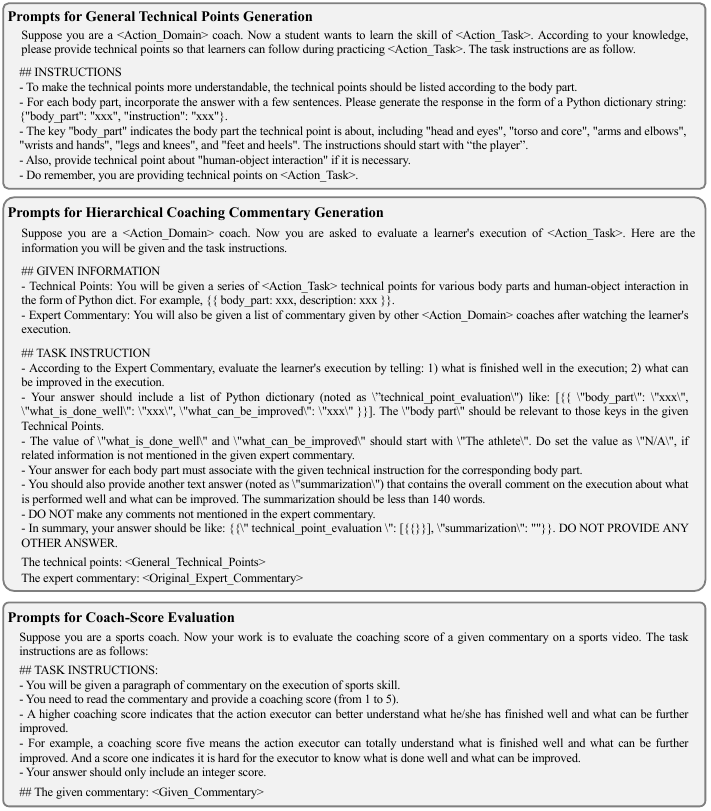}
    \vspace{-0.7cm}
    \caption{\textbf{The detailed prompts for generating General TechPoints (Upper), Hierarchical Coaching Commentary (Middle), and Coach-Scores (Lower).} ``\textless Action\_Domain\textgreater'': The action domain of the instance (\eg, Basketball, Soccer, Roch Climbing). ``\textless Action\_Domain\textgreater'': The action task of the instance (\eg, Reverse Layup). ``\textless General\_TechPoints\textgreater'': The generated general TechPoints for a specific action task. ``\textless Original\_Expert\_Commentary\textgreater'': The original expert commentary on the instance given by EgoExo4D \cite{egoexo4d}. Zoom in for best view.}
    \label{fig:prompts_dataset}
    \vspace{-0.3cm}
\end{figure*}

\vspace{0.1cm}
\noindent{\textbf{- Detailed Prompt for Annotation Collection}}. As described in Sec. \ym{3.2} of the main paper, we use LLM (GPT-4o) to generate general TechPoints and hierarchical coaching commentary. Additionally, as mentioned in Sec. \ym{3.3}, we prompt the LLM (GPT-4o) to compute the Coach-Score for commentary provided by various datasets. The detailed prompts are illustrated in \cref{fig:prompts_dataset}.

\vspace{0.1cm}
\noindent{\textbf{- Commentary length and vocabulary size}}. As shown in \cref{tab:len_vocab_size}, the vocabulary size of EE4D-DescCoach dataset is 7297. The TechPoint-level and Instance-level commentary spans average lengths of 15.8 and 68.9, respectively.

\begin{table}[t]
    \scriptsize
    \centering
    \scriptsize
    \resizebox{0.9\linewidth}{!}{
        \begin{NiceTabular}{l|cc}
        \toprule
        \textbf{Commentary} & \textbf{Length (Mean$\pm$Std)} & \textbf{Vocabulary size}\\
        \midrule
        TechPoint-level & 15.8$\pm$4.33 & 7297\\
        Instance-level & 68.9$\pm$14.1 & 7297\\
        \bottomrule        
        \end{NiceTabular}
    }
    \vspace{-0.3cm}
    \caption{\textbf{Lengths and vocabulary size of EE4D-DescCoach dataset}.}
    \vspace{-0.3cm}
    \label{tab:len_vocab_size}
\end{table}

\vspace{0.1cm}
\noindent{\textbf{Manual Check on Annotations}}. After obtaining the annotations with LLM, we manually check the evaluation set to ensure the annotation quality with the help of self-developed visualization tool.
We filter the misaligned instances or prompt the LLM to re-generate the annotation until it meets our requirements.

\vspace{-0.1cm} 
\section{Limitations and Future Work}
\vspace{-0.1cm} 
\label{sec:appendix_dicussion}

\noindent\textbf{- Scaling up TechCoach.} Our TechCoach is built with a relatively small text generator (136M).
We conduct fair comparisons with task-specific models training on our dataset with the same text generator and visual backbone (InternVideo2), and provide details analysis to show the contribution of the proposed Context-aware TechPoint Reasoner.
However, the small text generator with well large-scale pretraining may bring shortcomings on more difficult scenarios such as out-of-domain instances. Note that action coaching in OOD scenario is still an open problem in the related field. 
Due to the resource and computation constraint, we currently not able to integrate an LLM (7B or larger) with our TechCoach and train the Context-aware TechPoint Reasoner from scratch. 
We believe scaling up TechCoach, and conducting large-scale pre- and post-training will further enhance the DescCoach performance in various scenarios. We leave them for future work.

\vspace{0.1cm}
\noindent\textbf{- Hierarchical TechPoints.} In this work, for a given action task, we assume that each body part (or the dimension of human-object interaction) is associated with one paragraph of TechPoints. While this assumption provides a general framework for the initial exploration of TechPoint-aware DescCoach, TechPoints could be further constructed hierarchically.
For instance, when performing Mid-range Shooting, the shooting hand and guide hand play distinct roles in determining shooting accuracy and stability. Separating the TechPoint for high-level body parts (\eg, \emph{wrists \& hands}) into finer sub-parts (\eg, \emph{shooting hand} and \emph{guide hand}) could further guide the model to focus on fine-grained and specific aspects in an action execution.

\vspace{0.1cm}
\noindent\textbf{- Pose-Assisted DescCoach.} In this work, we focus on video-based DescCoach. While the Ego-Exo input enriches the visual context, there are cases where the egocentric video lacks sufficient visual information, and the human subject in the exocentric video appears relatively small. A valuable auxiliary modality to address this limitation is body pose. Since the original EgoExo4D \cite{egoexo4d} dataset provides frame-aligned 3D body pose data, incorporating pose sequences into the DescCoach process represents a promising direction for future research.

\newpage
\begin{table*}[t]
    \scriptsize
    \centering
    \scriptsize
    \resizebox{0.95\linewidth}{!}{
        \begin{NiceTabular}{p{2.8cm}|p{11cm}}
        \toprule
        \textbf{Technical Dimensions} & \textbf{General Technical Points}\\
        \midrule
        Head \& Eyes & The player should keep the head up and eyes focused on the basket to maintain good court awareness and accuracy. \\
        \midrule  
        Torso \& Core & The player should engage the core to maintain balance and stability throughout the movement, allowing for smooth and controlled layups. \\
        \midrule
        Arms \& Elbows & The player should use the arms to alternate the ball from one hand to the other while keeping the elbows close to the body to ensure control and precision during the layup. \\
        \midrule
        Wrists \& Hands & The player should use the wrists to flick the ball off the backboard and into the basket, ensuring a soft touch with the hands for better control. \\
        \midrule
        Legs \& Knees & The player should bend the knees to generate power for jumping, using the legs to propel themselves upward for the layup. \\
        \midrule
        Feet \& Heels & The player should position the feet shoulder-width apart for balance, pushing off the heels to jump and land softly to maintain control. \\
        \midrule
        Human-Object Interaction & The player should practice alternating layups from one side of the basket to the other, using the backboard as a guide to ensure the ball goes into the basket consistently.\\
        \bottomrule        
        \end{NiceTabular}
    }
    \vspace{-0.3cm}
    \caption{\textbf{General Technical Points for \emph{Basketball Drills - Mikan Layup}.}}
    \vspace{-0.2cm}
    \label{tab:all_TKP_mikan_layup}
\end{table*}

\begin{table*}[t]
    \scriptsize
    \centering
    \scriptsize
    \resizebox{0.95\linewidth}{!}{
        \begin{NiceTabular}{p{2.8cm}|p{11cm}}
        \toprule
        \textbf{Technical Dimensions} & \textbf{General Technical Points}\\
        \midrule
        Head \& Eyes                & The climber should keep the head up and eyes focused on the next hold. Always plan your next few moves in advance and be aware of your surroundings to maintain good spatial awareness.\\
        \midrule  
        Torso \& Core               & The climber should engage the core to maintain balance and stability. A strong core helps in keeping the body close to the wall, reducing the effort needed from the arms. \\
        \midrule
        Arms \& Elbows              & The climber should keep the arms slightly bent rather than fully extended to prevent fatigue. Use the arms to pull the body upwards and keep movements controlled and deliberate.\\
        \midrule
        Wrists \& Hands             & The climber should use an open hand grip when possible, conserving energy. Make sure to grip holds firmly and use your fingers to feel for secure positions on the rock. \\
        \midrule
        Legs \& Knees               & The climber should use the legs to push the body up the wall, which helps conserve energy in the arms. Keep the knees slightly bent to absorb movements and maintain flexibility. \\
        \midrule
        Feet \& Heels               & The climber should place the feet carefully on holds, using the toes for precision. Heels can be used to hook onto holds for added stability when necessary.\\
        \midrule
        Human-Object Interaction    & The climber should always check the stability of holds before putting the full weight on them. Use both hands and feet to distribute weight evenly and move smoothly from one hold to the next.\\
        \bottomrule        
        \end{NiceTabular}
    }
    \vspace{-0.3cm}
    \caption{\textbf{General Technical Points for \emph{Rock Climbing}.}}
    \vspace{-0.2cm}
    \label{tab:all_TKP_climbing}
\end{table*}

\begin{table*}[t]
    \scriptsize
    \centering
    \scriptsize
    \resizebox{0.95\linewidth}{!}{
        \begin{NiceTabular}{p{2.8cm}|p{11cm}}
        \toprule
        \textbf{Technical Dimensions} & \textbf{General Technical Points}\\
        \midrule
        Head \& Eyes                & The player should keep the head up and eyes focused on the ball at all times. This helps in tracking the ball's movement and anticipating its next position.\\
        \midrule  
        Torso \& Core               & The player should engage the core muscles to maintain balance and stability. A strong core helps in controlling the body's movements and maintaining proper posture. \\
        \midrule
        Arms \& Elbows              & The player should keep the arms slightly bent and relaxed at the sides to help maintain balance. The arms can be used to make minor adjustments to the body's position.\\
        \midrule
        Wrists \& Hands             & The player should use the hands to help with balance but should avoid using them to touch the ball. The wrists should be relaxed to allow natural movement. \\
        \midrule
        Legs \& Knees               & The player should keep the knees slightly bent to absorb the impact of the ball and to help with quick adjustments in positioning. This flexibility is crucial for maintaining control. \\
        \midrule
        Feet \& Heels               & The player should use the top of the feet to juggle the ball, aiming to keep it in the air with gentle, controlled touches. The heels should be lifted slightly off the ground to allow for quick movements.\\
        \midrule
        Human-Object Interaction    & The player should focus on making consistent, light touches with the ball to keep it in the air. Each touch should be calculated to keep the ball at a manageable height and within close proximity to the body.\\
        \bottomrule        
        \end{NiceTabular}
    }
    \vspace{-0.3cm}
    \caption{\textbf{General Technical Points for \emph{Soccer Drills - Juggling}.}}
    \vspace{-0.2cm}
    \label{tab:all_TKP_juggling}
\end{table*}

\end{document}